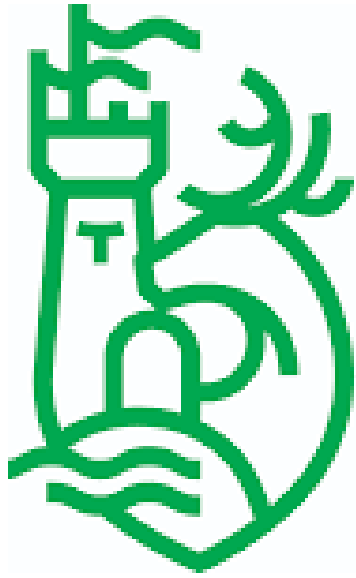

# Finetuning Transformer Models to Build ASAG System

Mithun Thakkar

23rd August 2021

Student ID: 20017138

Supervisors: Arash Joorabchi, Abbirah Ahmed

Masters of Artificial Intelligence and Machine Learning

Department of Electronic and Computer Engineering

# Abstract

Research towards creating systems for automatic grading of student answers to quiz and exam questions in educational settings has been ongoing since 1966 (Burrows *et al.* 2015). Over the years, the problem was divided into many categories. Among them, grading text answers were divided into short answer grading, and essay grading. The goal of this work was to develop an ML-based short answer grading system.

I hence built a system which uses finetuning on Roberta Large Model pretrained on STS benchmark dataset and have also created an interface to show the production readiness of the system. I evaluated the performance of the system on the Mohler extended dataset (2011) and SciEntsBank Dataset (converted). The developed system achieved a Pearson's Correlation of 0.82 and RMSE of 0.7 on the Mohler Dataset which beats the state-of-the-art performance on this dataset (correlation of 0.805 and RMSE of 0.793). Additionally, Pearson's Correlation of 0.79 and RMSE of 0.56 was achieved on the SciEntsBank Dataset, which only reconfirms the robustness of the system.

A few observations during achieving these results included usage of batch size of 1 produced better results than using batch size of 16/32 and using huber loss as loss function performed well on this regression task. The system was tried and tested on train and validation splits using various random seeds and still has been tweaked to achieve a minimum of 0.76 of correlation and 15% RMSE on any dataset.

The code for the system has been saved at https://github.com/mithunthakkar26/NLP-Projects

# CONTENTS

List of Tables

**List of Figures**

# 1 INTRODUCTION AND REPORT OUTLINE

## 1.1 SHORT ANSWERS

Short answers are natural language answers given to a question in free text. Fill-the-gap, short answers and essays are all considered as natural language answers which are a subset of recall answers, a terminology supported by literature (Jordan, cited in Burrows *et al.* 2015) and (Gay, cited in Burrows et al. 2015).

Even though short answers, fill-the-gap and essays are adjacent answer types, there are some key differences. Three aspects are considered to distinguish these three answer types as detailed by (Burrows *et al.* 2015) : length, focus, and openness.

- Length – fill-the-gap answers don't require more than few words while short answers and essays can be longer. Short answers can be small phrases to a maximum length of three to four sentences (Siddiqi et al., cited in Burrows et al., 2015). Essays however range from two paragraphs to several pages.
- Focus - short answers focus on content while essays focus on style (Shermis, Burstein, G¨utl, P´erez-Mar´, cited in Burrows et al., 2015).
- Openness - essays are considered more open-ended or subjective while short answers are objective or closed-ended.(Leacock and Chodorow, Siddiqi and Harrison, Wood et al., cited in Burrows et al., 2015)

The positioning of short answers in the hierarchical view of common question types as proposed by (Burrows *et al.* 2015) is shown in Fig 1.1. Reading comprehension is an exception in the diagram and does not form a part of the definition of short answers in context of this dissertation.

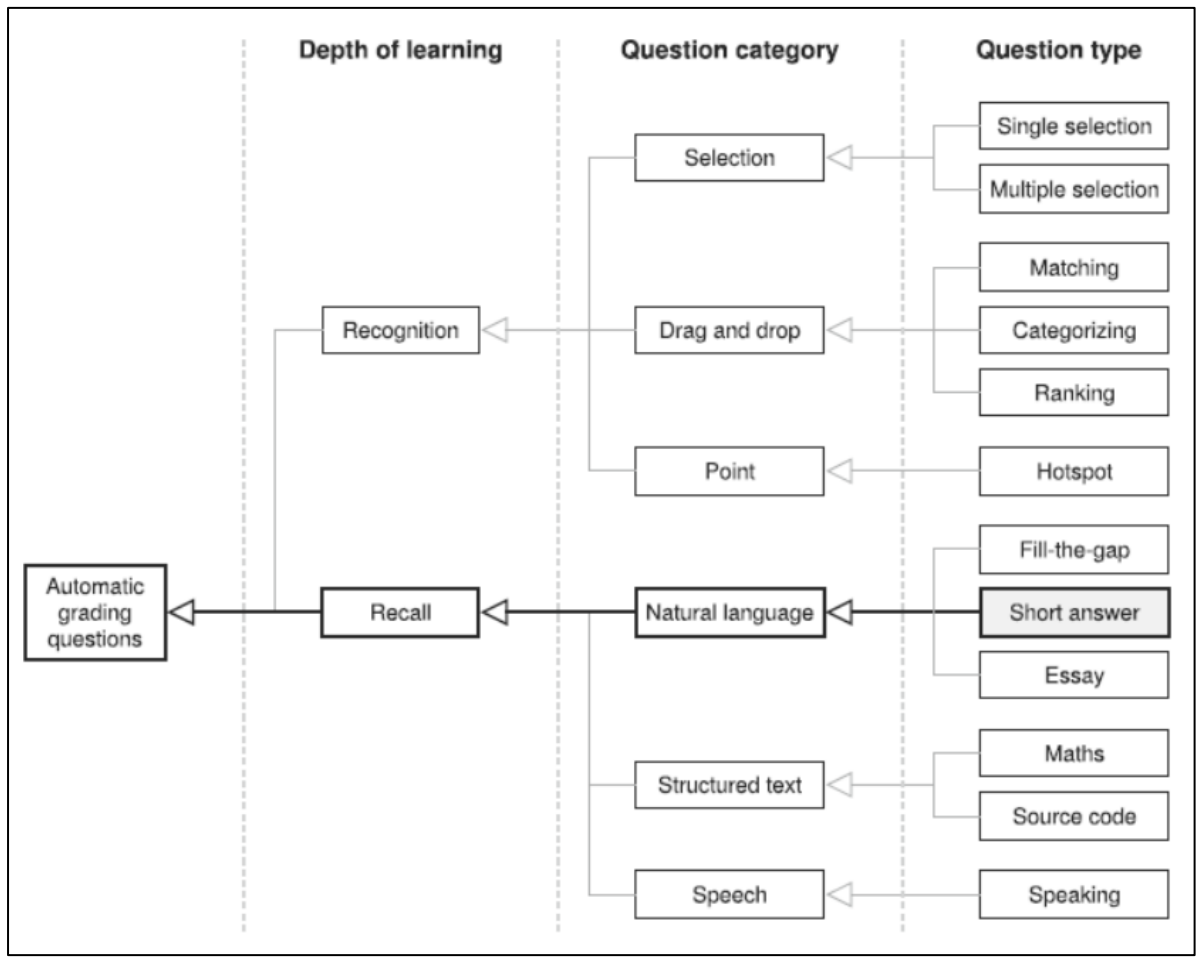

**FIGURE 1 A HIERARCHICAL VIEW OF COMMON QUESTION TYPES WHERE AUTOMATIC GRADING METHODS CAN BE APPLIED. SOURCE: (BURROWS ET AL., 2015)**

## 1.2 APPLICATIONS OF AUTOMATIC SHORT ANSWER GRADING (ASAG) SYSTEMS.

An ASAG system involves using automatic methods for grading short answers given by students. Benefits of using ASAG is threefold as listed below (Galhardi and Brancher 2018):

1. They can help reduce professors workload for grading students and allows professors to focus towards other tasks.
2. They can help provide an unbiased grading based only on student answers and not on any other external reasons. This would make grading more formalized.
3. They can provide quick results to students rather than make students wait until the manually generated results are out.

## 1.3 RESEARCH GOALS

- Explore use of word embeddings from a variety of pretrained transfer learning models such as BERT, SciBERT, RoBERTa to find appropriate vector representations for student answers and professor answers. These vector representations are used to find the similarity between the student and reference answers using string similarity metrics such as cosine similarity.
- Explore fine-tuning of transfer learning models to maximize correlation and minimize RMSE score as these two metrics are tracked for evaluating ASAG systems in literature.
- Explore all other techniques that would allow to improve performance of such a system
- Create a system which can be generally used for all ASAG tasks.
- Upon developing the system, I plan to publish a paper on this with the help of my supervisors.

# 2 LITERATURE REVIEW

## 2.1 POPULAR PUBLICLY AVAILABLE DATASETS USED TO CREATE ASAG SYSTEM

There are six popular publicly available datasets for the task of ASAG (Galhardi and Brancher 2018) viz. Texas dataset (Mohler *et al.* 2011), CREE and CREG datasets from (Meurers *et al.* 2011a), ASAP dataset from a Kaggle competion by (Hewlett 2012), Beetle and SciEntsBank datasets (Dzikovska et al., cited in Galhardi & Brancher, 2018).

Texas Dataset (Mohler 2011) consists of 87 questions from ten assignments with 4 to 7 questions each and two exams with 10 questions each relating to an introductory computer science module offered to undergrad students (Mohler *et al.* 2011). It has 2,273 student answers. Answers were graded on a scale of 0 to 5 by two human graders. The average grades of the two graders are provided in the dataset. Grader one is more generous in grading than grader two. The average grade given by grader one is 4.43 while that of grader two is 3.94. This dataset is also called as Texas Extended Dataset because it was an extension to the dataset that the authors created in 2009 (Mohler and Mihalcea 2009)..

The CREE dataset which stands for Corpus of Reading Comprehension Exercises in English comprises of in total 566 responses by English as a Secondary Language (ESL) students (Meurers *et al.* 2011a). These are responses to short-answer comprehension questions which were a part of an open book test. Collection of these responses were done from two different set of students at the same level having different teachers. The first set included 311 responses from 11 students who answered 47 questions and the second set have 255 responses from 15 students answering 28 questions. The first set is considered development set and second set is considered test set in (Meurers *et al.* 2011a).

The CREG stands for Corpus of Reading comprehension exercises in German. It is a dataset that was created from two German classes at Kansas University and Ohio State university (Meurers *et al.* 2011b). The Kansas University dataset consists of 117 questions and 136 target answers. It has 610 student answers given by 141 students with average number of tokens of 9.71. The Ohio state university dataset comprises of 60 questions with 87 target answers. It has 422 student answers given by 175 students. It has 15 tokens on average.

The 2012 Automated Student Assessment Prize (ASAP) from Kaggle comprises ten datasets relating to science, biology, and English Language Arts (Hewlett 2012). These are responses from grade 10 students and each dataset comprises an average of 2,200 individual responses consisting of one or a few sentences. It was scored by two graders on a scale of 0 to 2 or 0 to 3 depending on the dataset. Training dataset includes five fields: A unique essay ID, Essay set ID between values 1 to 10, Score given by grader one, score given by grader two and student response in ascii text.

The Beetle and SciEntsBank are a part of the Student Response Analysis corpus (Dzikovska *et al.* 2016). The beetle dataset is extracted from student interaction with Beetle 2 tutorial dialogue system (Dzikovska, cited in Dzikovska *et al.* 2016). It includes one to two sentence answers to 56 questions in the basic electricity and electronics domain. It has around 3000 student answers to the questions. The SciEntsBank dataset is from a corpus of student answers to assessment questions collected by (Nielsen *et al.* 2008). It has 197 assessment questions and 10,000 student answers in 15 different science domains. Each student answer in this dataset is labelled as Correct, Partially Correct/Incomplete, Contradictory, Irrelevant and Non-domain. The dataset has three distinct test sets – Unseen Questions (UQ), Unseen Answers (UA) and Unseen Domain (UD). Unseen Answers is a test set that comprises questions that are there in the training set but the student answers are not the same as that in the training set. Unseen Questions comprises questions that are not there in the training set. Unseen Domains is a domain independent set which was created by setting aside the complete set of questions and answers from three science modules from the fifteen modules in the SciEntsBank Dataset.

## 2.2 METRICS USED TO EVALUATE PERFORMANCE OF AN ASAG SYSTEM

For regression tasks, i.e. wherein the score is a continuous variable, Pearson's Correlation and Root Mean Squared Error (RMSE) are popular metrics. For classification tasks, mean quadratically weighted Kappa metric, mean F1 Scores, weighted F1 Scores and accuracy are popular metrics for evaluation.

Since this work would be focusing on arriving at a score i.e. a continuous variable, we would be using correlation and RMSE as evaluation metrics.

Formulas for these metrics are as below:

Correlation:

$$\rho(X, Y) = \frac{cov(X, Y)}{\sigma_X \sigma_Y}$$

where cov(X,Y) is the covariance between X and Y and Sigma (σ) is the standard deviation.

In our case, X is the actual value and Y is the predicted value.

RMSE:

$$\sqrt{\frac{\sum_{i=1}^{N}(x_i - \hat{x}_i)}{N}}$$

where RMSE is root mean squared error, i stands for the variable, N is number of data points, $x_i$ are the actual observations and $\hat{x}_i$ are the predicted values.

## 2.3 RELATED WORK (PAST 10 YEARS)

There have been variety of techniques used to create ASAG. No one technique has worked universally for all datasets nor have datasets been uniform. There has been a good variation in datasets in terms of domains and formats of datasets.

Several new techniques, innovations and discoveries have happened in the last 10 years that has boosted the power of ASAG systems over time. Below I would go about discussing on the research in this area that has happened in the last 10 years.

(Cutrone and Chang 2011) proposes using a component-based system utilizing a Text Pre-processing phase and a word/synonym matching phase for ASAG task. The system was not applied on a dataset for it to be evaluated and hence the paper does not have any numerical results to discuss about although it discusses on an evaluation plan to be implemented in the future. The method to follow is to use WordNet processing which involves usage of matching SharpNLP POS tags, word match and identification of positioning of words. For the system to work, there were required to be no misspellings which does not reflect real world conditions. It focused on getting single sentence responses which also is impractical. It required text tagging and punctuation removal. NGrams detection was used. It required removal of stop words. Stop words can include negation as well as other important words which may add semantic value to a sentence. Removing stop words can lead to potential loss of semantics.

In (Klein *et al.* 2011), LSA is used in extracting and representing the contextual usage meaning of words by statistical computations applied to a large collection of data. The system was tested using Pearson's Correlation metric and was applied on seven questions from 2007 first year computer science algorithms exams which had 282 in total submission. The system achieved more than 0.8 correlation for all questions except one which was a good achievement for the time the system was created. This system however required generating ideal configurations for each question for it to work. This is nontrivial in nature and impractical in real world scenario. This system suffers with the problem of not performing well in case of less training data. Vanilla LSA has also been known to ignore syntactic and grammatical structure of text as mentioned in (Omran *et al.* 2014). Despite the drawbacks which this technique posed at the time when it was created, today's modern NLP techniques use the basic idea of LSA i.e. representing words in vector space to derive semantic as well as contextual representations.

(Mohler *et al.* 2011) proposes usage of features based on dependency graphs and word similarities for calculating distance between student and model answers. This is then used in machine learning techniques to automatically score student answers. To evaluate the performance of their system, they created their own dataset (Mohler 2011). The best scores achieved by their system included Pearson correlation of 0.464, RMSE of 0.978 and Median RMSE of 0.862. The paper also had introduced the concept of question demoting which was adopted by research in several years to come. Question demoting involves removal of stop words that have already appeared in questions from reference and student answers.

In (Gomaa and Fahmy 2012), the authors tested separately and combined 13 String and 4 Corpus based similarity measures as features to a regression model to achieve maximum correlation value of 0.504 on the Mohler Dataset 2011 (Mohler *et al.* 2011). String based similarity measures included character-based distance measures and term-based distance measures. Corpus based similarity measures involved usage of latent semantic analysis, Explicit Semantic analysis, pointwise mutual information and distributional similarity. The system involved removing stop words and stemming as a part of preprocessing stage. This is followed by creating a similarity matrix and finding overall similarity. Even though this paper proposes very simple techniques for ASAG, it makes one very big assumption that words with similar meaning occur in similar context. This assumption has been addressed with by papers that come in the later years with creation of contextualized word vectors.

(Omran and Ab Aziz 2013) proposed two methods – the automatic sentence generator method and usage of three algorithms – Common Words, Longest Common Subsequence and Semantic Distance. Using automatic sentence generator method involves producing an alternative model answer using a synonym dictionary. Even though this method involves making use of semantic aspects of words in the answers, this can be highly inefficient in terms of computing as it involves storage of large number of words. This method also does not consider the contextual meaning of words (instead it only considers the synonyms). They used a dataset which included 40 questions in each of three assignments answered by three students i.e. 360 short answers. They found that their system achieved Pearson's correlation of 0.82. Even though the system achieved good results, it was tested on a relatively small dataset and testing it on a larger dataset would bring more conclusive results.

This paper also talks about traditional methods for tokenization which involve removal of white spaces and symbols. Current day NLP techniques involve usage of symbols too as they do add to the semantic meaning of the entire sentence. It also discusses the removal of stop words. The CW, LCS and SD algorithms proposed for ASAG are too brute force techniques. The first two techniques do not even consider the semantics of the words while SD does not consider the context.

(Basu *et al.* 2013) proposed to use a combination of unsupervised approach of clustering along with professor grading to attempt ASAG. It involves clustering similar student answers using training a classifier to identify if the student answers should be clustered. This would allow the professor to grade clusters of student answers together. This approach is different from previous approaches in the sense that it addresses the question of misunderstandings of concepts by students. By having well defined clusters of students – the professor can address these misunderstandings well. It also focuses on giving grades as well as feedbacks to students based on which clusters they belong to. The authors used a subset of responses from the United States Citizenship Exam which included in total 798 responses. They evaluated the system by identifying how far a grader could get to with the given amount of effort. Results showed that their method required 63% fewer actions than the LDA based approach of clustering and 36% fewer actions than the metric classifier operating on individual items. Although the approach followed in this paper is thorough and even achieves good performance in clustering answers, it is a bit complex for the end user to understand as is highlighted in the paper. It can take a lot of active discussion between the machine learning team and the users for successful implementation of such system.

(Gomaa and Fahmy 2014) compared 14 string similarity and two corpus-based measures for finding semantic similarity between answers and used K-means clustering to cluster marks into categories and given feedback. They introduced a new Arabic dataset that contains 50 questions and 600 student answers. The correlation and RMSE scores by combining the best metrics (in terms of correlation and RMSE) were 0.854 and 0.8 respectively. These were achieved by using a machine learning approach.

(Brooks *et al.* 2014) created a interface for enabling clustering based ASAG to be used in real world scenarios. They used the clustering approach with 25 teachers and compared the performance of teachers in grading the assignments with and without clustering based system. Each teacher would grade 200 distinct answers from 2 questions in an online environment. The 2 questions were from the Powergrading dataset (Basu *et al.* 2013) and the clustering algorithm used was also from the same paper. Based on their survey with the teachers, they found that more number of teachers preferred clustered interface more than the unclustered interface in all parameters. While the system helped reduce teacher effort, performance can be affected when there are wider range of responses from students. This could lead to more number of cluster formations.

(Jang *et al.* 2014) creates an ASAG for Korean Short answers. They used a concept-based scoring method that can evaluate complex questions. They used morphological analysis to build a token-based scoring template for improving the coverage of their process. The dataset consisted of 7 questions and 21,000 student responses from 2012 NAEA. They used these many student responses to prove that their system can be used for large scale responses. Results showed that error rate was very low and agreement between grader and system were 95%.

(Kulkarni *et al.* 2014) combines short answer grading with peer assessment wherein the grading algorithm predicts a grade with confidence on the grade assessed. The grading algorithm was a generic text classifier using etcml.com with predicted grade as output. It uses text features like word, bi-gram and tri-gram counts, length of answers and letter n-grams. Based on the confidence achieved from the system and the rubric, the student graders grade the answers. The system was evaluated with 1570 students in a large online class. They found that by using this technique, accuracy of grading improved. This technique obviously performed good because it was heavily assisted with human effort which is not desirable in a robust ASAG system.

(Pado and Kiefer 2015) dives into creation of domain independent system for ASAG as a complement to manual grading using sorting of similarity metric of reference and student answer. For evaluation of the system, they used the CREG dataset and introduce a new corpus which comprises computer science short answers in German. The similarity between student and reference answer is calculated using a technique of DKPro Similarity implementation of Greedy String Tiling (Wise 1996, cited in Pado and Kiefer 2015. Their system achieved the state-of-the-art performance on CREG and slight drop in performance on their German corpus. I believe using just this one technique is a very handsoff approach to the ASAG problem. There are several similarity metrics that have been used in the past by Gomaa and Fammy which could have been used in this paper.

(Sultan *et al.* 2015) discussed using an unsupervised and a supervised system which take in a pair of sentences and outputs a value to signify the amount of similarity between the sentences. The unsupervised system considered the word alignments while the supervised system considered the word alignments and sentence vector (word2vec) similarities (cosine) of the two sentences. They used the SemEval 2015 dataset(Agirre *et al.* 2015) and STS test data(Agirre *et al.* 2012), (Agirre *et al.* 2013) and (Agirre *et al.* 2014) from 2012 to 2014 for evaluation. Their supervised system achieved best correlation of 0.8015 with human annotation on the STS dataset while their unsupervised system achieved correlation of 0.7919. This paper was a general paper addressing not only ASAG task but also other tasks involving sentence similarity.

(Zehner *et al.* 2016) used clustering for clustering student answers based on their numeric representations and then applied supervised ML algorithm on top to assign an interpretable code. To assign the code, they used within cluster class distribution and Bayes theorem, which is represented as probability that response k, belonging to jth cluster g has been assigned to ith class c by the external criterion. They used the German PISA sample dataset which includes samples from 15-year-olds and ninth graders from Germany. There were between 4152 and 4234 test takers. Agreements between human and system were of 76% to 98% in terms of Fleiss Kappa for within computer reliability across repetitions.

(Zhang *et al.* 2016) used answer-based, questions-based and Student Models for feature engineering and supplies the features to six classifiers including deep belief networks, SVM and Naïve Bayes. They used similarity metrics for student model features like max-matched idf, length difference, cosine similarity and weighted text similarity. For question model features, they used Q-matrix which represents the relationship between individual questions and KCs. They also used question difficulty for question model. They used the Cordillera training corpus (Carolyn 2007 and VanLehn et al. 2007, cited in Zhang et al. 2016). They used accuracy, AUC, Precision, Recall, and F-Measures for evaluation. Results show that their feature engineering models outperformed (achieved more than 0.75 score in all metrics) conventional answer-based models and among the six classifiers, DBN performed best. In my opinion, having separate features for questions makes a lot of sense as students should uniformly perform better in some questions rather than others. The paper also discusses on using student models. I believe using student models is not practical as student behaviors can change and tracking a multitude of students is inefficient. I hence think that student model features do not add much value to prediction.

(Sultan *et al.* 2016b) was a paper written specifically focusing short answer grading as compared to Sultan's previous year paper which focused on semantic similarity in general. Text similarity, question demoting, term weighting and length ratio features are used for the Mohler dataset (Mohler *et al.* 2011) and SCIENTSBANK corpus (Nielsen *et al.* 2008). The Mohler dataset was used for regression task while SCIENTSBANK was used for classification. This paper achieved mediocre results on the Mohler dataset with correlation of just 0.63 and RMSE of 0.85 (in domain training) which were state of the art on this dataset at the time. F1 scores on the SemEval 2013 datasets was mediocre (~0.55) too. Yet, this paper improved upon the previous results on the datasets and does add value to the ASAG research in general.

(Magooda *et al.* 2016) focused on using vector-based techniques for ASAG. Preprocessing steps included stop words removal. Seven word-to-word similarity measures were used viz. Block Distance, JiangConrath, Lesk, DISCO, Word2Vec, GloVe, Sense Aware Vectors. These were applied on four datasets viz. Extended Texas Dataset (Mohler *et al.* 2011), Texas Dataset (Mohler and Mihalcea 2009), Cairo university Dataset (Gomaa and Fahmy 2014) and SemEval 2013 dataset (Dzikovska *et al.* 2013). Three Sentence-to Sentence similarity measures were used viz. Text-to-text model, vector summation model and Min-Max Additive model. On the Mohler 2011 dataset on which this dissertation is on, the performance was a rather mediocre one as the best scores in paper displayed RMSE of 0.659 and correlation of 0.586.

(S. Roy *et al.* 2016) discuss an unsupervised approach using sequential pattern mining and a scoring process they have created. They use the Texas Dataset (Mohler and Mihalcea 2009), Extended Texas Dataset (Mohler *et al.* 2011) and a dataset they created on reading comprehension assignment for standard 12 students. They achieved Pearson's correlation of 0.57 on Texas Dataset, 0.67 on Extended Texas Dataset and 0.41 on the reading comprehension dataset they created. Even though the approach achieved comparable results with other approaches, and it is an unsupervised approach, it did so on the assumption of wisdom of students.

(Kumar *et al.* 2017) use Siamese Bidirectional LSTMs, a pooling layer based on Earth Mover Distance and a regression layer for the ASAG task. They apply these techniques on the Texas Dataset and SemEval Dataset (SciEntsBank). They achieve Pearson's correlation of 0.649 and RMSE of 0.83 on the Texas Dataset. On the SemEval Dataset, they achieve correlation and RMSE of 0.554, 0.157 and 0.237 and 0.758, 0.996 and 0.958 on Unseen Answers, unseen questions, and unseen domain, respectively. The SemEval dataset has categorical labels. To report correlation figures, the categorical labels were converted into ordinal labels. Data Augmentation was applied on Texas dataset as it has 2442 data points. In modern day machine learning, with the creation of efficient transfer learning models, importance of data augmentation has reduced as one can train even on smaller datasets – this is called as training on a downstream task.

(Riordan *et al.* 2017) explores how neural techniques used in Automatic Essay scoring apply to the task of ASAG. They use three datasets – ASAP-SAS (Hewlett 2012), the Powergrading dataset (Basu *et al.* 2013) and the SRA dataset (Dzikovska *et al.* 2012). They experimented with different parameters like mean-over-time layer (Taghipour and Ng, cited in Riorden et al. 2017), pretrained embeddings, features from a convolutional layer, network representational capacity, bidirectional architectures, regression vs classification and attention layer. On the ASAP-SAS dataset, they used 250-dimensional bidirectional LSTM and attention mechanism. On Powergrading, they used CNN features with window length 5, 150-dimensional bidirectional LSTM and attention mechanism. On SRA, they use a 300-dim unidirectional LSTM with attention mechanism. Based on their experimentation, they found – that finetuning pretrained embeddings helped in majority of the datasets, neural models benefit from similar size hidden dimensions, Mean-over-time produced competitive results on many prompts, bidirectional LSTMs and attention produced the then best results. Even though this research paper did not present conclusive results as it wasn't its objective, it does help in getting insight into what one can expect while using neural techniques including LSTMs for the ASAG task.

(Jayashankar and Sridaran 2017) discusses a model which uses semantics with visual appeal in form of word cloud to partially automate free text evaluation. It helps in evaluation by making use of relative and cohesion cloud. Preprocessing involved removal of numbers, punctuation, whitespaces and stop words. This is followed by synonym handling using WordNet, substituting plurals and building term document matrix. The output then generates cohesion and comparative word cloud. The dataset used consisted of questions and answers from IGCSE syllabi for grade 10 by 13 students. Total questions were 4209 across 8 modules. The model generated score standard deviation from the mean of 2.81.

(Hassan *et al.* 2018) used paragraph embeddings of student and reference answers and found cosine similarity between them. The cosine similarity was then used as a feature for a regression model to predict the score. There were two approaches used for generating vector representations – one was by using the sum of word vectors for the words in the answer and second approach involved training a deep learning model to infer the paragraph vector of a given answer. Word2Vec, Glove, Fasttext and Elmo were used to get word embeddings. Doc2Vec, InferSent and Skip-Thoughts were used to get paragraph embeddings. The Texas extended dataset was used for this paper. Ridge Regression was applied on the generated features. The lowest RMSE and correlation coefficient achieved were 0.797 and 0.569 respectively using Doc2Vec. In my opinion, even though the paper did not show any state-of-the-art performance, it does help us understand the performance of different paragraph embeddings on the data.

(Galhardi *et al.* 2018) introduces a new Portuguese dataset involving 13 teachers, 12 undergraduate students, 15 questions and 245 elementary school students. This means there were in total 3675 (245*15) answers in the dataset. Grades were given ordinally in range of 0 to 3. The inter-rater agreement obtained by the 12 undergraduate students was 0.581. It also discusses an automatic short answer grading system for this dataset. The method includes 4 sets of features – bag of ngrams, lexical similarity, semantic similarity and other features like length ratio, word length average etc. On average, best accuracy achieved for 4-class classification across the 15 questions was of 69%. This performance was achieved by Ngrams.

(Horbach and Pinkal 2018) proposes semi-supervised clustering. The paper tries to improve on previously published paper results that were achieved from clustering from employing an unsupervised approach. In previous approaches, manually labelled data was used only for post clustering label propagation while in this paper, human annotation is set to be before and during the clustering process. Human Annotation was used for feature selection (because clustering approaches suffer from noisy features), for setting constraints and for centroid based label propagation. Two datasets were used – ASAP and PG. Performance achieved in this paper from using MPCKM clustering (with pairwise CL constraints and metric learning) was better than clustering approaches back then and came closer to supervised approaches. Kappa achieved was 0.566 as against previous clustering approaches kappa of 0.504. Given that clustering approaches have already reached perfection in prior studies, I believe this paper brings a new dimension to clustering by introducing a semi-supervised approach.

(Saha *et al.* 2018) observes that prior techniques using just sentence embeddings performed well on in-domain while dint do well on out-of-domain circumstances. The paper hence introduces a new feature encoding using Word overlap, histograms of partial similarity and Histogram of partial similarity with POS tags and Question types. Using the combination of sentence embeddings and this new feature encoding produced better results. It acknowledges that in dialog-based evaluation scenario, students mention tokens in a non-sentential manner which might not get captured by semantic similarity or textual entailment approaches. The experiments were done on SemEval Dataset (SciEntsbank), extended Texas dataset and Large-Scale Industry dataset (a psychology dataset with correct, partial and incorrect labels). Results achieved from applying the techniques achieved better results as compared to the then state-of-the-art performance.

(Pribadi *et al.* 2018) raises two questions on sentence similarity measures approach – whether the method accounts for variety in student answers in the reference answers and whether the sentence similarity measure is accurate. To resolve the first question, it proposes using MMR method and for second question, it proposes using GAN-LCS. MMR (Maximum Marginal Relevance) is used to generate reference answer variants from student answers. GAN-LCS is an extension of LCS (longest common subsequence) method. A part of GAN-LCS is the similarity coefficient which can deal with finding similarity between sentences of significantly different lengths. The coefficient value is then converted into Student grade. These Techniques were applied on the Extended Texas dataset. The correlation and RMSE achieved on the dataset were 0.468 and 0.884 respectively. The paper concludes by mentioning that the proposed method is done without training and without the use of corpus. I believe having pretraining approaches which make uses of corpus even though appear expensive are more practical than the proposed approach. That's because they generate real semantic and contextual representations of sentences.

(Zhang *et al.* 2019) focuses on specifically semi-open-ended questions which may require subjective responses from students. They propose using CBOW model for generating features in form of word vectors using domain-general information and domain-specific information. Domain-general information is taken from Wikipedia while the domain-specific information is the labelled student answers. LSTM was used to encode word sequence information for building the classifier. QWKappa was used to evaluate the agreement among the grades. Results show that using domain-specific as well as domain-general information produces better results rather than using either of the two on standalone basis. Modern day NLP techniques like BERT have evolved from the ideas presented in the paper and inherently account for domain-specific and domain-general information using transfer learning. It also accounts for sequence information using position embeddings.

(Gomaa and Fahmy 2019) propose using Ans2vec (Skip-thought vector approach) to convert reference and student answers into vectors. The authors consider skip-thought vectors as sentence version of word2vec i.e. word2vec predicts surrounding words while skip-thought vectors predict surrounding sentences. At first, the authors generate skip-though vectors of student answers and reference answers using pretrained sentence embeddings. They then compute component wise product and their absolute difference and predict the score using log linear classifier on top of the extracted features. Three datasets were used to experiment this system: Texas extended dataset, Cairo University Dataset and SciEntsBank Dataset from SemEval. On the Texas dataset, Correlation of 0.63 and RMSE of .91 was achieved. On the Cairo dataset, correlation of 0.79 and RMSE of 0.92 was achieved. F1 Results for SciEntsBank were 0.58, 0.47 and 0.46 for UA, UQ and UD respectively. In my opinion, the Texas Dataset has been consistently difficult to work with as most research hasn't been able to achieve more than 0.7 of correlation on the dataset. Same can be seen in this research.

(T. Liu *et al.* 2019) proposes a system which uses transformer blocks and multiway attention layer (instead of single attention channel) to capture relation between reference and student answers. The system was applied on K-12 dataset which contains 120,000 pairs of student and reference answers from an online education platform. AUC and accuracy were used to evaluate performance of the system. Accuracy of 0.89 and AUC of 0.94 was achieved. In my opinion, training your own dataset using transformers would only work for purely online tutoring involving large student numbers. This wont work in majority of scenarios where the datasets aren't that large and, in this situation, employing transfer learning can be computationally more efficient and generally applicable.

(Surya *et al.* 2019) compares various deep learning specific models for ASAG. They compare CharCNN, CNN, LSTM and BERT on the ASAP – SAS dataset. For evaluation, they use Quadratic weighted kappa metric. BERT achieves best performance with Mean QWK across all prompts to be 0.71. Although they arrived at the conclusion that BERT performed best, they observed poor performance on test. I believe that should be because of using standard parameters as given in the BERT paper.

(Sahu and Bhowmick 2019) used traditional text similarity-based features like WordNet similarity, latent similarity analysis and others and introduced new features like topic models and relevance feedback-based features. They built a stacked ensemble-based model using a combination of different regression models. They tested their system on the Texas extended dataset, the SciEntsBank and Beetle Dataset. For feature extraction, semantic similarity features, lexical overlap features, information retrieval measures, topical similarity features, relevance feedback-based features and alignment-based features were used. For grading models, individual and ensemble models were both used, and the authors arrived at the conclusion that stacked ensemble-based learning gave better performance. The proposed system achieved pearson correlation of 0.703 and RMSE of 0.793 on the Mohler dataset. It also achieved state of the art results on the SciEntsBank and Beetle corpus in terms of F1 scores. In my opinion, the proposed system is efficient and well thought out. It is not that computationally expensive as the deep learning approaches and produces good results.

(Sadr and Nazari Solimandarabi 2019) compares various measures of semantic relatedness and similarity for ASAG. They also apply an approach which uses students' answers with the highest score as feedback to improve precision of semantic relatedness. The semantic relatedness measures used were corpus-based and knowledge-based measures. They use a dataset which contains 630 student answers. Answers are individually scored by two annotators in the range of zero to five. The correlation between the scores assigned by two human judges is 0.7228. They conducted three sets of experiments. In the first part, the performance of corpus-based and knowledge based semantic relatedness measures is evaluated. In the second part, the effect of domain and size of background is explored. The effect of using the using the students' answers with highest scores to improve precision was explored in the third part. In the first experiment, among all the measures used for semantic relatedness, ESA achieved the highest Pearson's correlation of 0.513. In the second experiment, ESA with using Wikipedia as the background achieved 0.513 correlation. After applying pseudo feedback in the third part, LSA performed best which achieved a correlation of 0.545.

(Sung *et al.* 2019a) explores and evaluates as research question as to whether updating pre-trained BERT LM whilst using unsupervised domain corpora in updating LM for SAG is helpful in improving ASAG performance. They find out that it does impact the performance positively. Second research question they explore is as to how generalizable the effect on unseen domains is and find out that it is not generalizable. Another research question for the authors was as to how labeled question-answer data can be exploited to update LM for short answer grading. For this question, they understood and advise on using Question answering corpora for LM update in addition to fine-tuning. While the first three research questions seem quite obviously answerable from current practices in using BERT with the use of Huggingface models, the fourth question and its results were interesting. BERT along with QA corpora for LM update did achieve better results as compared to BERT alone.

(Sung *et al.* 2019b) used BERT finetuning for SciEntsBank Dataset and achieved 10% absolute improvement in macro-averaged F1 score on the then state-of-the-art results. They experimented with increasing the number of epochs for finetuning, tried understanding the finetuning impact on cross-domain performance. They also checked their systems efficiency on two psychology domain datasets. They achieved Accuracy of 75.9, 65.3 and 63.8 on Unseen answers (UA), unseen questions (UQ) and unseen domains (UD) respectively with using learning rate of 2e-5 for SciEntsBank. Mean F1 were 72, 57.5 and 57.9 for UA, UQ and UD respectively. Weighted F1 were 75.8, 64.8 and 63.4 for UA, UQ and UD respectively. They used learning rate of 3e-5 for the psychology dataset and achieved good results on that dataset too. For the rest, they used standard setup for BERT with 4 epochs, and batch size of 32. Increasing batch size to 12 did not provide any better results. They tried to identify the effectiveness of transfer learning and effectiveness for data-starved problems and found that the approach they followed was good at both. They also however found that their approach did not perform well on cross-domain data. After reading this paper, it seems that finetuning BERT can help achieve perfect results. Although, I believe that finetuning was easily done on this dataset. That is not the case with other datasets.

(Basak *et al.* 2019) used several matching rules based on recognizing entailment relation between dependency structures of two answers. The method followed involved pre-processing, dependency parsing, lexical alignment, categorization of tokens, graph matching and assignment of scores and calculation of the grade. The dataset used for this method was the Texas extended dataset. The best correlation score achieved was of 0.365 using DISCO trained on English Wikipedia corpus. I believe the correlation score achieved is too low in comparison to the then state of the art.

(Süzen *et al.* 2020) used the Texas Extended Dataset to solve the problem of ASAG using Clustering, hamming distance and regression analysis. They used the vocabulary of the student answers to cluster their responses and found correlation between the scores given by the teachers the clusters. They found the correlation between the clusters and the scores to be strong. They used hamming distance between the model answer and student answer and found that the correlation between their computed hamming distance and the scores assigned to be high. Using the hamming distance as feature, they arrived at a correlation of 0.81 and 0.83 for the two human graders. The MSE of the linear regression was 0.09 and 0.1 for the two human markers. This paper was too lengthy with too many basics about the ASAG task explained with relatively lesser explanation on technicalities involved. Additionally, one very big drawback that the authors acknowledge too was that synonyms or semantics or order of words were not accounted for in their approach.

(Tan *et al.* 2020) explore the use of two-layer graph convolutional network to encode student responses from the heterogenous graph created. Sentence level and bigram level nodes are a part of the graph. Based on the co-occurrence relationship, an edge is constructed between the nodes. The edge weights which log the correlation degree are calculated from the sentence-level TF-IDF value or the PMI value. The SemEval dataset SciEntsBank was used for this paper along with a two subject junior school Chinese dataset. The two subjects here are a maths dataset and a literature dataset. Data augmentation was applied on the datasets because training a deep learning model requires large amount of data to train. They found that on all the three test sets Unseen answers, unseen questions, and unseen domains, they believe their system outperformed baselines – even though in most cases – BERT performed almost equally well. On the 'All' 5-way dataset of the SciEntsBank which includes UA, UQ and UD, they achieved best Accuracy, Mean F1, and weighted F1 of 44.6, 44.3 and 42.7 respectively while BERT achieved 42.8, 43 and 41.7 respectively. In my opinion, the approach that is followed in this paper is difficult and computationally expensive as compared to BERT – given that BERT achieves similar results. BERT does not require data augmentation efforts or any sort of pre-processing and can be used very efficiently with even small datasets.

(Condor 2020) trained BERT on expert ratings of constructed responses and use an additional automated grading system for calculating Cohen's Kappa to find agreement of system with human rater. When the system's reliability metric does not have a high enough score, the author's automated model calls attention to ratings where a second opinion might be needed. The author uses the DT-Grade dataset which consists of short constructed answers from tutorial dialogues between student and an intelligent tutoring system. It comprises 1100 student responses from 34 distinct questions relating Newtonian Physics. The system achieved testing accuracy of 0.76 and Cohen's Kappa of 0.684. One thing which is different about this paper as compared to previous works is that the author has experimented with different setups in terms of epochs, batch size and learning rate. He has not plainly followed the setup that has been provided in the BERT paper. This helped him achieve better results.

(Ghavidel *et al.* 2020) used BERT and XLNET for ASAG on the SciEntsBank dataset 2-way, 3-way and 5-way tasks. They used BERT-Base and XLNET-Base for their experiments. They trained 10 epochs with dropout probability of 0.1, warmup proportion of 0.1, batch size of 16, learning rate of 5e-6 for BERT and 5e-5 for XLNET. In the 2-way task, BERT base uncased achieved state of the art results for Unseen answers. In 3-way, BERT base uncased achieved state of the art performance on unseen questions in Accuracy and Weighted F1 scores and in terms of accuracy for unseen domains. In 3-way, XLNET achieved state of the art performance on mean F1 and Weighted F1 for unseen domains. In 5-way, both BERT and XLNET achieved better performance than priorly used models for this dataset. It isn't a coincidence that these models were explored for the ASAG task in the same year as MPNET (Song *et al.* 2020) was introduced. MPNET harnesses the power of XLNET in considering dependencies between predicted tokens and BERT in considering the full position information of a sentence. MPNET would perform better on the ASAG task because of these reasons.

(Filighera *et al.* 2020) was written to test the robustness of ASAG Systems which generated state of the art results back then. The author specifically mentions results generated in the paper (Sung *et al.* 2019b). By prepending short token sequences to the student answers, they could trigger a way to artificially improve grade assigned. In their experiments, they found out that using this method, students could pass an exam which has a passing threshold of 50%. They found that on the unseen answers split in SciEntsBank dataset, prepending the trigger sequences "none varies" and "none would" managed to increase misclassifying by 8.8% of samples from incorrect answer to correct answer. On the unseen question split, the sequence "none would" lead to increase in misclassification of 10.1% of samples. On the "none elsewhere" increased misclassification by 12.1%. They also found that the trigger length, usage of objective functions and transferability did impact misclassifications. Even though the work in this paper sounds not towards the constructive side of things, it does throw light on how critical it is to build a robust system. It also throws light on how the impact of psychological semantics can play a role in ASAG systems because of contextual and semantic representations of word vectors.

(Zhang *et al.* 2020) discusses in the abstract that deep learning models have not been applied to ASAG. This suggests that the authors did not research enough to know that deep learning models have been extensively applied to ASAG. The results from the paper showed that the performance of ASAG improved upon using student and question models and Deep belief networks can be applied to ASAG.

(Camus and Filighera 2020) was yet another excellent paper by Filighera after (Filighera *et al.* 2020). It captures the practices followed in today's NLP industry with the usage of Huggingface library. They finetune different transformer-based architectures on the SemEval 2013 dataset. They found that Roberta based models performed better than other models, multilingual models did generalize better across languages than their English counterparts, transfer learning a model trained on MNLI performed better when finetuned on the SemEval dataset and knowledge distillation was suitable for ASAG specific tasks in case of lower GPU resources. This paper really helped me with my work in the remaining part of the dissertation. I will refer to it when I discuss on my own work in this dissertation.

(Lun *et al.* 2020) propose using multiple data augmentation strategies along with finetuning BERT for ASAG systems. These comprise of back-translation, correct answer as reference answer and swap content. Back-translation involves translating the entire sentence from one language to another and from that language back to the original language for data augmentation. This approach is better rather than just using synonyms of words from Wordnet. This helps in maintaining syntax, semantic and contextual information. In the correct answer as reference answer technique, the student answer which was given a perfect score was considered as a reference answer and compared against the student answers which got imperfect score. The swap content strategy involves first using the back-translation strategy to generate twin sentences and simply generating combinations of reference questions, reference answers and student answers. There can be 2^3 combinations of such twin sequences. For BERT finetuning, they used BERT-BASE, and followed the same configuration approach as (Devlin *et al.* 2018). They used the SemEval SciEntsBank dataset for the task and performed their experiments on unseen answers (UA) split alone. They evaluated the performance using accuracy, Mean F1 and Weighted F1. They achieved better results using data augmentation on all the three - 2-way, 3-way and 5-way grading in the UA split of the dataset. This paper really answers the burning question of how to augment data for semantically and contextually smarter architectures like BERT and other pretrained architectures.

(Kishaan *et al.* 2020) propose using active learning to assist in the task of grading short answers. They built a web-based GUI for an interactive interface to enable this system. The system which uses machine learning for ASAG queries the human grader for grades of the data samples it is most uncertain about and there by reduces time and effort of graders in assessment. They used three datasets – the Texas Extended

Dataset, SemEval 2013 Task 7 dataset and an in-house ASAG dataset. To evaluate the performance of the system, they used number of clicks made by the grader to grade answers. Results from the paper show that with lesser number of training samples, the same level of performance was achieved with more samples whilst reducing time and effort of graders with reduced number of clicks. I believe such a system can be used in production upon implementing a stable ASAG system.

(Chaturvedi and Basak 2021) propose using three models to score student answers individually – the first model uses POS tags between reference answers (RA) and student answers (SA), the second model subjects the RA and SA individually to a parser which converts each of them into a set of dependency triples. The third model arranges words in RA and SA into two-dimensional matrix and similarity scores between the answers are found using Word2Vec and fastText. The average of the scores is used to assign the final score to the student. They achieved a high correlation score of 0.805 on the Texas extended dataset using the techniques. Even though they achieved high correlation, the method employed was too rule based. They used Word2Vec and fastText which are good models, but they don't take contextual representations of words into consideration.

(Tulu *et al.* 2021) proposes using MaLSTM and sense vectors obtained by SemSpace, a synset based sense embedding method leveraging WordNet. Synset representations of student and reference answers are inputted into a parallel LSTM architecture which leads to transformations of these representations into sentence vectors and Manhattan similarity is found between the vectors. They use this method on Texas Extended dataset and achieve a Pearson correlation of above 0.95 and RMSE of 0.04. While this may seem like the State of the Art, but they achieved this by taking each question separately and split answers into train and test split. Upon inputting all answers to all questions together, they achieved a correlation of 0.15 and MAE of 0.23. Even though the system performed well on individual questions, it did not perform well on aggregate dataset. This does not work well in practical implementations or reflect any real-world applications. Additionally, LSTMs do not consider long range dependencies as good as transformers.

# 3 ANALYTICAL BACKGROUND

As can be seen from the literature review, over the last 3 years, there has been an increase in usage of "deeper" learning-based approaches towards creating ASAG systems. This allowed for ASAG systems to not only consider semantic and syntactic representations of words but also contextual representation of words. A major reason behind that is the invention of the transformer (Vaswani *et al.* 2017).

This invention ignited the creation of robust deep learning models like BERT (Devlin *et al.* 2018), RoBERTa (Y. Liu *et al.* 2019), GPT(Radford *et al.* 2018), GPT2(Radford *et al.* 2019), GPT3(Brown *et al.* 2020), MPNET(Song *et al.* 2020), AlBERT(Lan *et al.* 2019), CodeBERT(Feng *et al.* 2020), SciBERT(Beltagy *et al.* 2019) etc.

Brief explanations of the models are as below:

- The transformer model is a simple network architecture based on attention mechanism which helps it to be more parallelizable than priorly used RNN, LSTM and CNN based approaches. Attention also allows for lower computational complexity per layer in the network and allows for long term dependencies to be learned better.
- BERT stands for Bidirectional Encoder Representations from Transformers can pretrain deep bidirectional representations from unlabeled text by jointly conditioning on left and right context in all layers. Hence, it could be finetuned to variety of downstream tasks including semantic similarity. It was pretrained on BooksCorpus and Wikipedia.
- The authors of RoBERTa established that BERT was very undertrained, and they modified BERT which improved performance on GLUE and SQUAD tasks. They pretrained BERT on bigger batches over more data, removed next sentence prediction objective, trained on longer sequences, and dynamically changed the masking pattern applied to the training data. They also added additional corpora to their training called the CC-News corpora.
- MPNET is pre-training model that leverages XLNET's(Yang *et al.* 2019) permuted language modelling whilst also taking auxiliary position information as input for making the model see the entire sentence and hence decreasing position discrepancy. This way it inherits the advantages of BERT and XLNET and ignores its disadvantages.
- ALBERT is a model which was introduced to reduce difficulties relating to limited GPU/TPU resources when training BERT. The authors provide two parameter lowering techniques to reduce memory consumption and improve training speed of BERT. They also make use of self-supervised loss which emphasizes modeling inter-sentence coherence. This particularly helps when downstream tasks have multi-sentence inputs. Because of this, ALBERT achieved then SOTA results on GLUE, RACE and SQuAD benchmarks.
- CodeBERT is a bimodal pre-trained model that learns general-purpose representations that support downstream natural language and programming language applications. It is trained using hybrid objective function that inculcates pretraining task of replaced token detection, that is to find plausible alternatives sampled from generators.

- SciBERT is a model based on BERT for training it on scientific data. It uses unsupervised pretraining on a big multi-domain corpus of scientific research for improving performance of downstream scientific tasks.
- GPT (stands for Generative Pre-training) is a unidirectional transformer model which was trained on BooksCorpus.
- GPT-2 is an evolution of GPT and was trained on the WebText Corpus. The largest GPT-2 model is a 1.5 B parameter Transformer that achieved SOTA results when it was released on 7 out of 8 tested language model results in zero-shot setting.
- GPT-3 is an autoregressive model with 175 B parameters and can be applied well for few shot learning and one shot learning tasks.

As a result of such deep learning-based architectures being used for NLP tasks, an ecosystem of open sourced models has been created on the website https://huggingface.co/models (Wolf *et al.* 2019). These include raw models as well as models that have been finetuned on a different task, which can be used for further finetuning on another downstream task. The advantages of finetuning this way in context of ASAG have been explored in a few research papers in recent years (as mentioned in the literature review).

# 4 SPECIFICATION AND DESIGN, IMPLEMENTATION AND TESTING

## 4.1 SPECIFICATION AND DESIGN

### 4.1.1 Technical Specifications

The system was built using the below specifications and it would work without any problems if these specifications are used. On an overall basis, the system is platform independent.

Python version: 3.6,

GPUs: Tesla V100-SXM2-16GB or Tesla P100-PCIE-16GB (would recommend using V100 for best performance),

Pytorch Version: 1.9.0+cu102

Storage Capacity: 10 GB

### 4.1.2 DESIGN

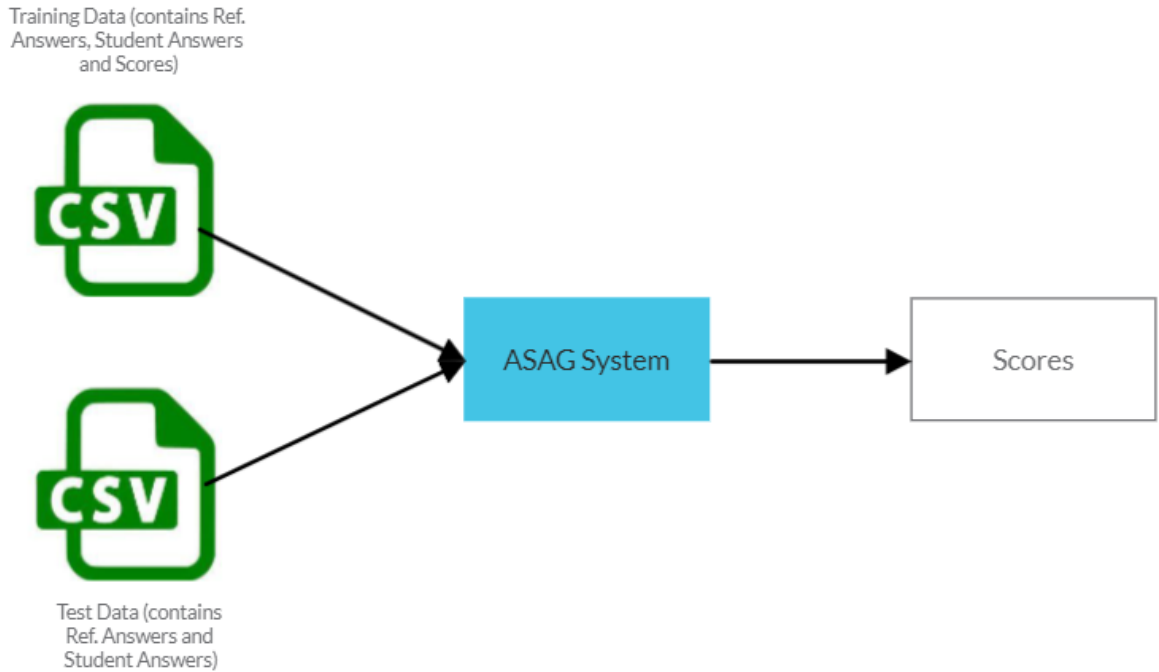

**FIGURE 2 A VERY HIGH-LEVEL VIEW OF THE SYSTEM (TRAINING FIRST TIME AND SCORING)**

Figure 2 shows the design of the system from the user's point of view. For training the system, the user uploads the training data csv file which contains reference answers, student answers and scores. Along with this, the user also uploads test data i.e. the answers which the user seeks to get checked. The system would check these answers and generate scores.

If the user has already trained the system on a particular dataset, he would have to only upload the test data file. This can be seen in Figure 3.

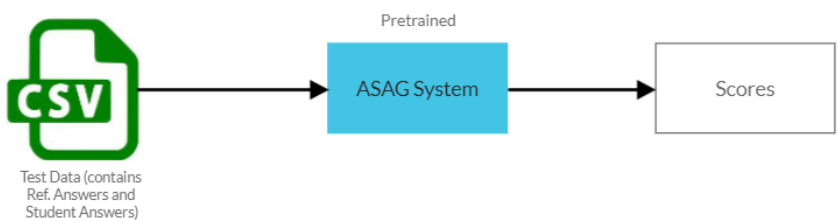

**FIGURE 3 A VERY HIGH-LEVEL VIEW OF THE SYSTEM (SUBSEQUENT SCORING)**

Below are the instructions on using the interface for training the data.

**FIGURE 4 STEP ONE OF USING THE INTERFACE**

**FIGURE 5 STEP TWO OF USING THE INTERFACE**

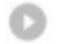

**FIGURE 6 STEP THREE OF USING THE INTERFACE**

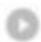

**FIGURE 7 STEP FOUR OF USING THE INTERFACE**

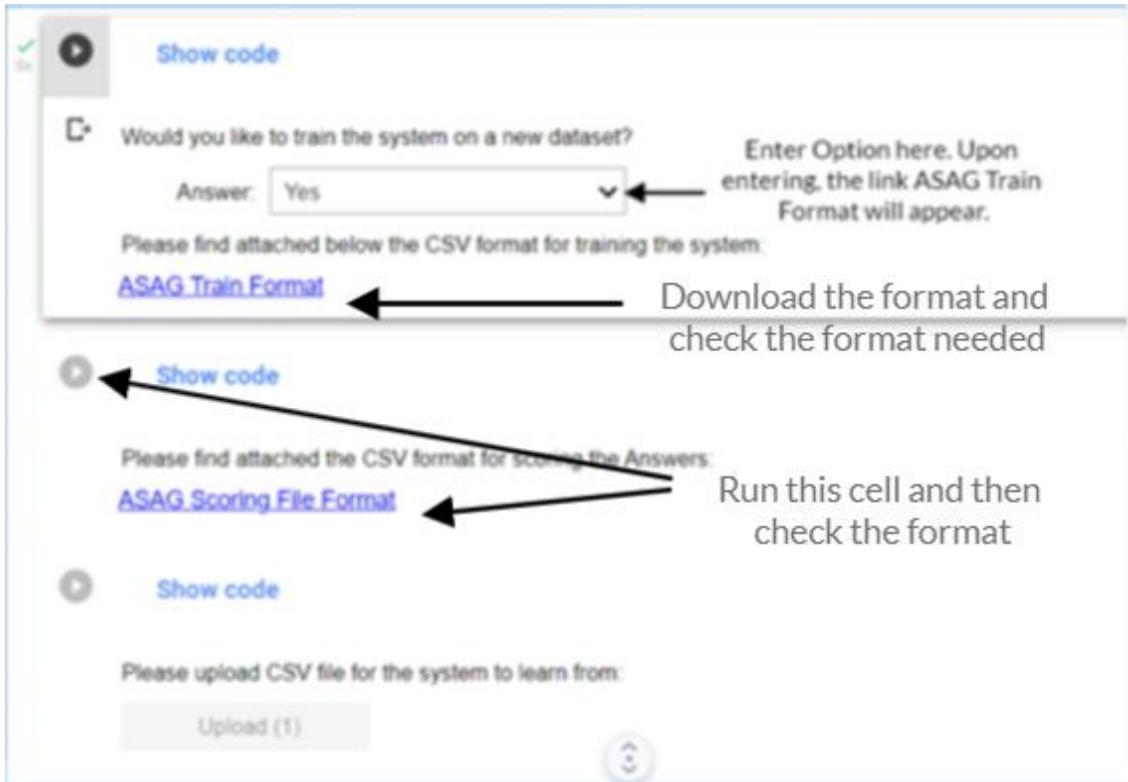

**FIGURE 8 STEP FIVE OF USING THE INTERFACE**

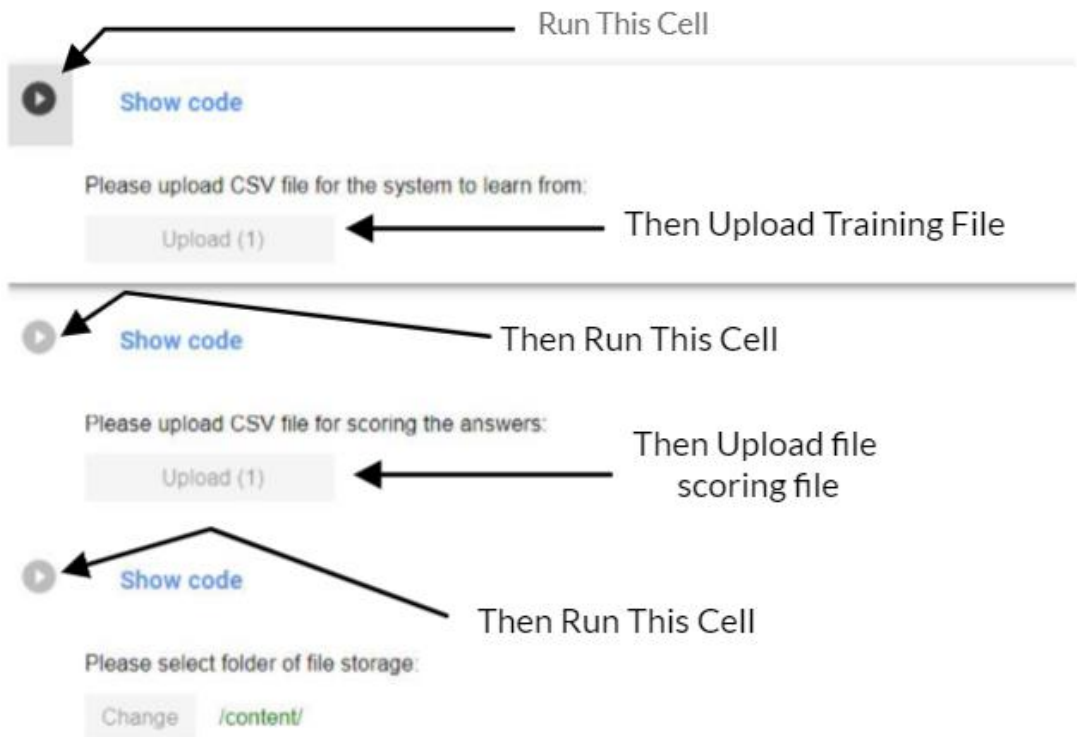

**FIGURE 9 STEP SIX OF USING THE INTERFACE**

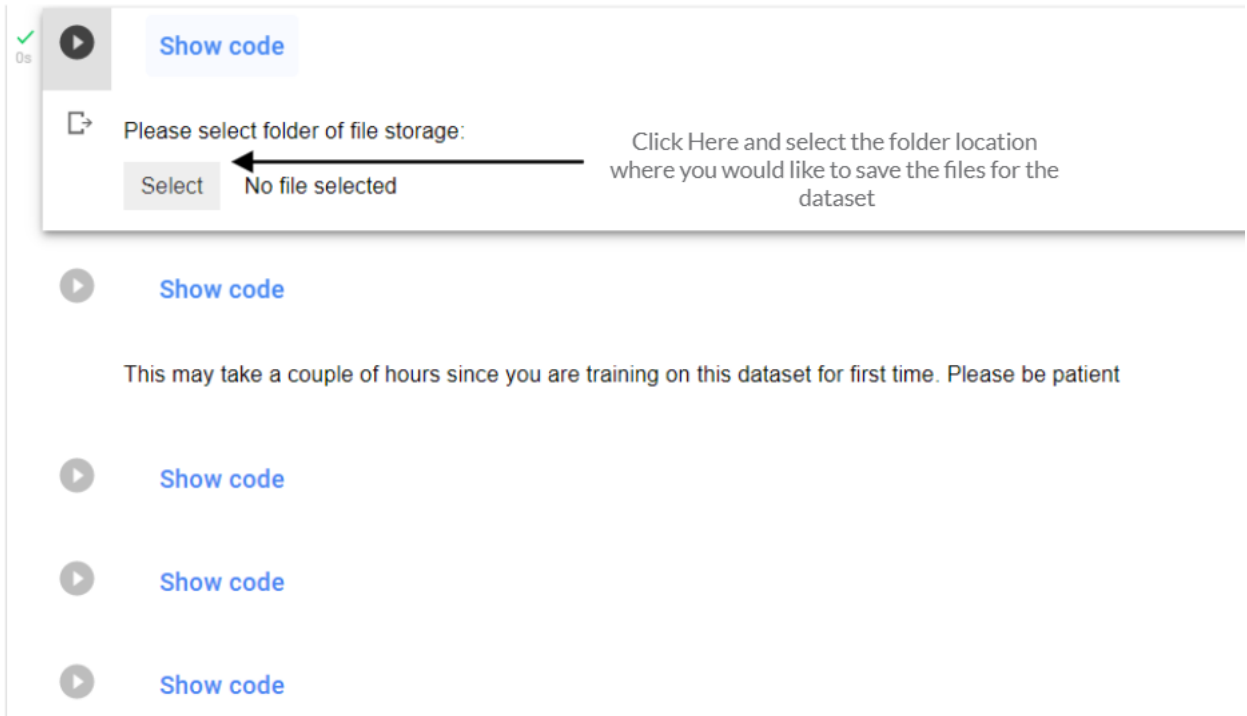

**FIGURE 10 STEP EIGHT OF USING THE INTERFACE**

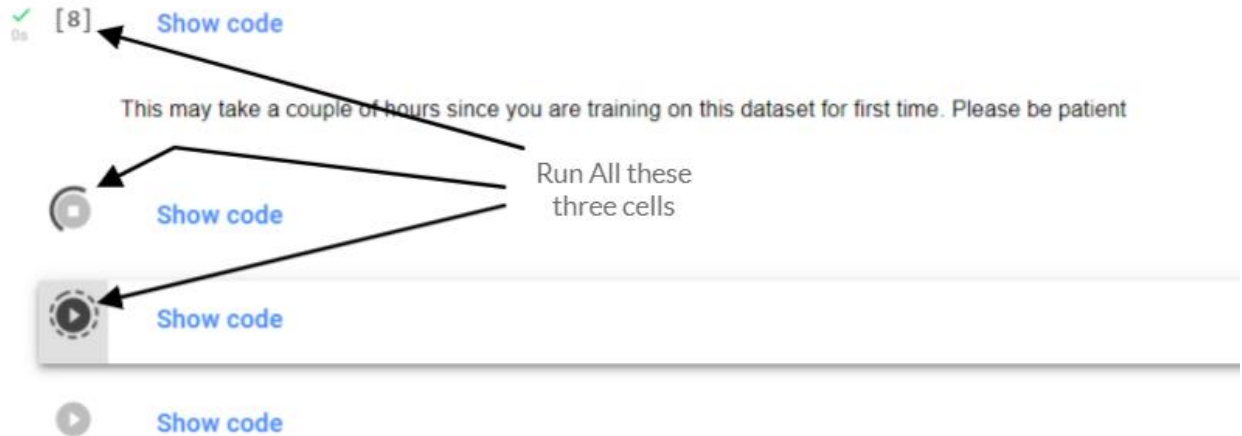

**FIGURE 11 STEP NINE OF USING THE INTERFACE**

**FIGURE 12 SCORES GENERATED IN TABULAR FORMAT**

**FIGURE 13 SCORES IN EXCEL AND PIVOT TABLE**

Above, the user enters the dataset name and answers whether he would like to train the system on a new dataset. If yes, he would upload the csv file for training the system and the test data file (without the scores). If no, he would upload only the test data file (without the scores) because the system won't need to be trained again as it already has been trained on that dataset. Scores for the test data file are output along with the answers for the user to see. The user gets a hyperlink which would allow him to download the csv file which contains the answers and the scores. Post this, a pivot table is generated which would allow the user (professor) to compute averages, minimum, maximum scores etc. on student answers to specific questions.

## 4.2 IMPLEMENTATION AND TESTING

### 4.2.1 IMPLEMENTATION

Below is the diagram from the BERT paper to demonstrate how the finetuning works. This implementation can be generalized to all transformer-based models.

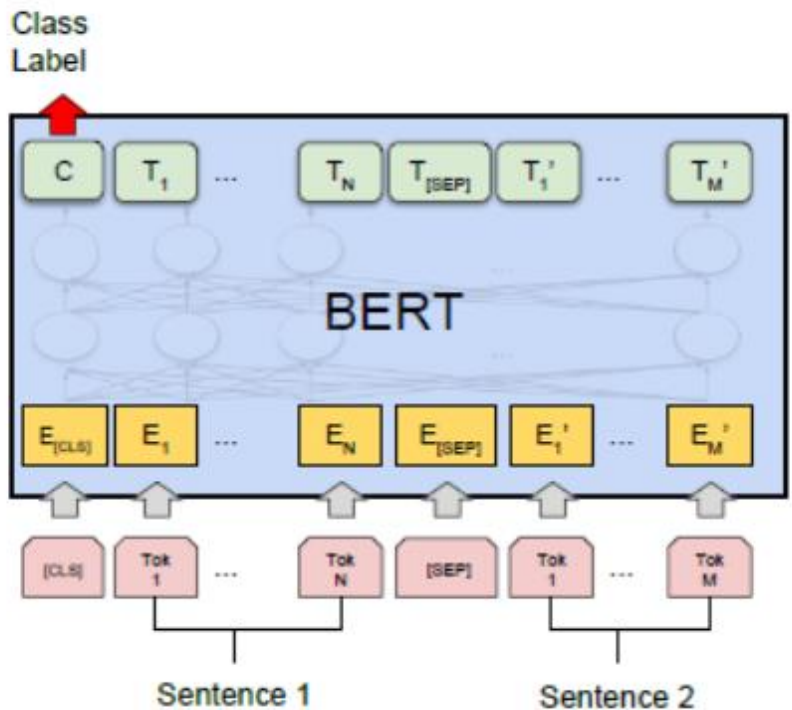

**FIGURE 14: GENERAL DESIGN OF TRANSFORMER MODEL FINETUNING FOR SENTENCE SIMILARITY TASKS (Devlin *et al.* 2018)**

As can be seen above, we feed in two sentences into a transformer-based model. These sentences are tokenized by the tokenizer for these models. For the model to identify these sentences as two separate sentences, we use a special token ([SEP] in case of BERT). The CLS token is also appended at the beginning of the input. These tokens as well as words of the sentences that are input are converted into numeric form. Padding or truncation is performed on the sequences based on the max length parameter mentioned by the user. Default max length is set to 512 for BERT i.e. at a time, the system can be inputted with 512 tokens comprising of two word sequences or sentences. These sequences can be more than one English sentence (not necessarily a single English Sentence). A sentence as referred to in the BERT paper means a word sequence and not an English sentence. In this project, I have set the max token length to 256 in total for the two sequences because this project is meant for short answers and to improve system speed performance. From the output above, we get a vector corresponding to the CLS token. If we are performing a regression task – which is in our case, we use linear layers to receive this vector and output a score. Even though in the above diagram, we can see BERT is mentioned, this general design is relevant to any transformer model. The special tokens may be different, but the design remains the same. In practice, the entire process of inputting the sequences is automated using the Autotokenizer class from HuggingFace library and model is also autorecognized by AutoModel class from HuggingFace.

There are three finetuning or transfer learning techniques that one can use for generating predictions from the pretrained models – training the entire architecture, training some layers while freezing the others and freezing the entire architecture. In this project, I initially froze the entire architecture to check the performance of different models for the datasets. I then unfroze the architecture for those models which gave best performance. I finally narrowed down to use STS Roberta Large Model (*cross-encoder/stsb-roberta-large)*.

I experimented with the below models from the huggingface library for performing finetuning on. From my experiments, I shortlisted models and their parameters based on how good learning occurred for these models on the dataset. I did this by looking at the learning curves and correlation and RMSE performance on validation and test sets.

- *roberta-large-mnli* : I used this model because it has been mentioned in (Camus and Filighera 2020) that Roberta models tend to generalize better on ASAG tasks. It was also mentioned that models trained on MNLI improve performance on the task.
- *cross-encoder/stsb-roberta-large*: This is a Roberta large model and has been trained on STS corpus (which stands for semantic textual similarity) using Sentence Transformers Cross Encoder Class (Reimers and Gurevych 2019).
- *cross-encoder/quora-roberta-large*: This model is also a Roberta large model and has been pretrained on Quora which has a community answering to questions including several domains including scientific domains.

- *allenai/scibert_scivocab_uncased*: This model is also relevant because it has been pretrained on scientific publications corpus.
- *microsoft/codebert-base*: This model is relevant because the Mohler extended dataset is relating to computer science and has computer science specific notations used in it.
- *sentence-transformers/stsb-mpnet-base-v2*: This MPNET model is pretrained on STS and inculcates advantages of BERT and XLNET but ignores their limitations.
- *albert-large-v2*: ALBERT models perform well on GLUE tasks as per ALBERT paper. Our task which is relating to sentence similarity is a GLUE task.
- *sentence-transformers/nli-mpnet-base-v2*: This MPNET model was finetuned using Stanford NLI dataset.

Using my experiments, I found that *cross-encoder/stsb-roberta-large* gave the best performance in terms of correlation and RMSE for two datasets (about which has been discussed in the Results section below). I achieved the results by using the below hyperparameters:

Learning Rate: 1e-5, weight decay: 0.1, dropout rate 0.1, warmup proportion: 0.06, batch size: 1. I used the below architecture on top of the Roberta Architecture to get estimates from:

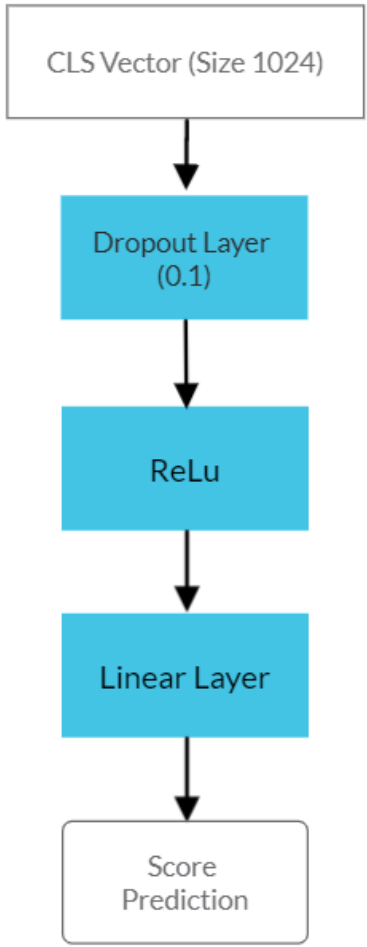

**FIGURE 15 FINETUNING LARGE TRANSFORMER ARCHITECTURE**

I performed early stopping by selecting the model with the lowest validation loss. The loss function I used was Huber Loss (Huber 1992) for arriving at the final results. The formula for this function is below:

$$L_\delta(y, f(x)) = \begin{cases} \frac{1}{2}(y - f(x))^2 & \text{for } |y - f(x)| \leq \delta, \\ \delta\left(|y - f(x)| - \frac{1}{2}\delta\right), & \text{otherwise}. \end{cases}$$

Below is the Huber loss function's curve in contrast to Squared error curve:

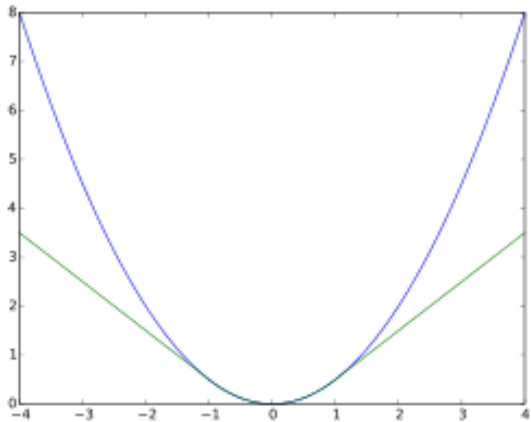

**FIGURE 16 HUBER LOSS FN (GREEN) AND SQUARED ERROR LOSS FN (BLUE)**

In **FIGURE** 16, X axis has the predictions and the Y axis has the loss values. As can be seen, the more extreme the predictions, the higher would be the loss value for squared error. That is relatively smoothened in case of Huber loss. One other advantage of using Huber Loss is that it is differentiable at zero because we see a smooth curve at prediction value zero above. This is in contrast to Mean Absolute Error (MAE), which is not differentiable at zero (on the MAE curve). It is for these reasons that Huber Loss is considered a better loss function as compared to MAE and RMSE.

I wasn't aware of this loss function initially and hence did not use it in my initial experiements. I had used RMSE loss in my initial experiments (which are discussed about in the Experiments section below).

### 4.2.2 TESTING

I have used the Mohler Extended Dataset and SciEntsBank Dataset from the SemEval competition for evaluating the robustness of the system.

The Mohler Extended Dataset consists of 2442 student and reference answer pairs. The Mohler dataset has a continuous output as score i.e. score is assigned in range of 0 to 5 (could have decimal values).

The SciEntsBank corpus is a well-organized corpus which has 4969 student-reference answer pairs in the training set, 540 pairs in the unseen answers test set, 733 pairs in the unseen questions test set and 4562 pairs in the unseen domains test set. Students for this dataset in 5-way system have been evaluated using the grades – contradictor, correct, irrelevant, non_domain and partially correct/incomplete.

The evaluation on both datasets involves finding Pearson's correlation and RMSE values as has been done in the literature.

I have split the Mohler dataset into train, test and validation sets (80-10-10) by stratification such that answers to each question are present in each of train, test and validation set. The validation and training sets were used for finetuning. These were split using random seed of 40. The final evaluation was then done on the test set. These were split using random seed of 43.

I have also evaluated instances where duplicate student-reference answer pairs would be at first removed and then the answers to the same question would be made available in train, test and validation sets. In a real-world scenario – students would give same short answers especially when the answers are as short as one or two words. For example - reference answer to the third question in 4th assignment of Mohler Dataset is "by reference" and several student answers to that question are answered as "by reference".

For the SciEntsBank Corpus, the authors themselves have provided with unseen answers, unseen questions and unseen domain test sets. The unseen answers (UA) test set comprises answers to same questions which are present in training set. Unseen questions (UQ) test set comprises answers to questions which are not present in the training set. Unseen domains (UD) test set comprises answers to questions from a completely different domain from that present in the training set. Students for this dataset in 5-way system have been evaluated using the grades – contradictory, correct, irrelevant, non-domain and partially correct/incomplete. These grades are not comparable with the Mohler Dataset because scores assigned in Mohler dataset are a continuous variable. Hence, I have reassigned these grades as score of 2 for correct, score 1 for partially correct/incomplete and score 0 for the remaining. For this idea, I have taken inspiration from the paper (Kumar *et al.* 2017).

Since the UA split of the SciEntsBank is most relevant for us for comparing to performance on Mohler dataset, I have focused on the performance achieved on the UA split in this project.

For finetuning, I split the training set of SciEntsBank dataset into 90-10 i.e. 90 for training and 10 for validation. These splits are stratified such that answers to the same questions are present in validation and training set.

The Mohler dataset is provided in the form of a directory of folders - annotations, docs, parses, raw, scores and sent. The sent folder consists of individual text files one for each question. It also comprises files all, all-sent, answers and questions. In the 'answers' and 'questions' files, there are reference answers and reference question respectively. In the 'all' file, there are answers from students. I have fetched the reference answers and student answers from the 'answers' and the 'all' files. The scores folder consists of scores assigned to the students. I have fetched these scores as well. I have put these three things into a dataframe and performed my computations on these.

The SciEntsBank dataset has been provided in a directory which comprises of the folders – reliability, test-unseen-answers, test-unseen-questions, test-unseen-domains and train. Each of folders (except reliability) has folders – Core, dependency and extra. The Core Folder comprises of .xml files which have the questionID, reference answers and student answers which I have loaded into a dataframes in Python using Pandas.

### 4.2.3 IMPORTANT POINT ABOUT SPLITTING THE DATASET FOR PRODUCTION READY RESULTS

In real world scenario, we may split the dataset into train and validation sets which have been carefully stratified to be representative sets of each other, yet there are many a times when validation set is not a good representation of the training set. It could also be possible that hyperparameters that work on one split of training set might not exactly work on another split of the training set. For example, in this project, when I randomly split the Mohler dataset into training, test and validation split, training was very good on a particular split. This was because the hyperparameters used were doing very well on that split. Upon using others splits, the training wasn't that good. I readjusted the hyperparameters for those splits and training was very well. Upon having good training, the test results were also favorable. This is a problem because this means the system does not work in circumstances where the dataset is not favorable (dependent variable has issues in it).

I found a workaround for this problem – I created a while loop which would rerun the system with a different random split of train and validation set if it did not produce a correlation of more than 0.75 on the validation set with the previous split. This sounds very time consuming, but it did work for me in practice. To reduce the time required for running the system, I added a condition in finetuning code which would check if RMSE is higher than 0.15 after six epochs, then the finetuning would abort. I kept six epochs based on my observations of the training process (i.e. the RMSE values during each epoch). I also considered the impact of warmup applied towards initial training and having a batch size of 1 which leads to rapid convergence. If RMSE does not reduce below 0.15 in first six epochs, the finetuning would be exited and restart with a new random split of training and validation set. The random split always would be stratified such that validation set, and training set would all have answers to same questions. It is because of this that the best result (of 0.82 correlation and 0.73 RMSE) on the Mohler Dataset is achieved on a favorable train validation split (random seed of 40) because the scores in this dataset are skewed. However, in the worst-case scenario, the system is configured to still achieve minimum correlation of 0.76 and RMSE of 0.73 which is still close to the state-of-the-art performance. One might cast a doubt on this technique, but I have a few reasons why this technique would work:

First reason is that I am discussing about splitting the training and validation set while not touching the test set while finetuning. Second reason is I was getting good results on the UA test set of SciEntsBank Dataset using the same hyperparameters as I did with Mohler Dataset (SciEntsBank Test Sets have been provided by the authors of the dataset and are widely accepted in the industry). third reason is that the Mohler Dataset is skewed with more students achieving higher scores than those achieving lower scores. Upon diving deep into the scores provided in the dataset, I saw that the answers for questions 12.7, 12.5, 9.7, 12.1, 9.2 and 1.4 were all but one given a score of 5 while answers for question 8.2 were all given a score of 5. Neural networks or any other machine learning models don't learn well when there are misclassifications.

## 5 RESULTS AND DISCUSSIONS

The system achieved Correlation of 0.82 and RMSE of 0.7 on the test set of Mohler Dataset. It achieved correlation of 0.79 and RMSE of 0.56 on the Unseen Answers split of SciEntsBank Dataset.

Below are the learning curves for them:

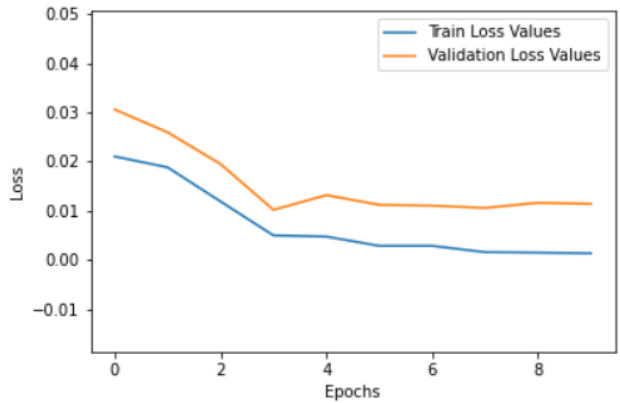

**FIGURE 17 FINAL RESULTS – LEARNING CURVES (MOHLER)**

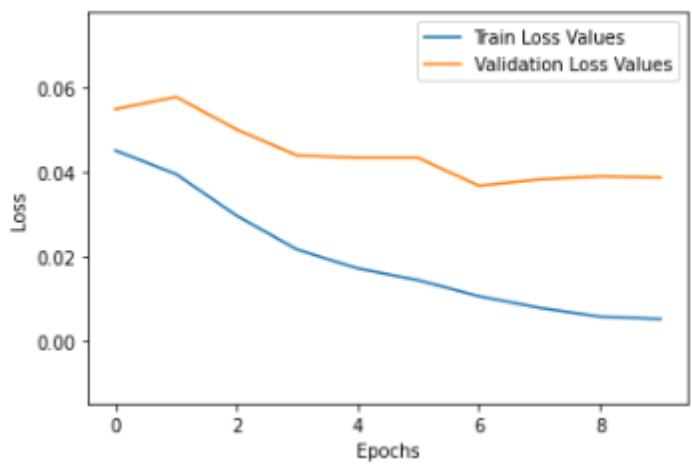

**FIGURE 18 FINAL RESULTS – LEARNING CURVES (SCIENTSBANK)**

Overall, I am very happy with the results as they show promise on two publicly available datasets which have acceptance in the research community. As can be seen in the curves, there is some divergence in the validation loss in the later epochs. For this reason, the code used in the system captures the epoch with the lowest validation loss value and saves the model in a checkpoint. In Deep Learning terms, I implemented Early Stopping.

These results were achieved by unfreezing the transformer architecture and would stubbornly not be achieved by freezing the architecture. They also showed improvement when loss function was changed to Huber Loss instead of RMSE loss. Since an unfreezed architecture is a very deep network with 12 encoder decoder stacks for base models and 24 encoder decoder stacks for large models, it makes sense to use heavy regularization. I hence applied regularization by minimally using just one linear layer, one ReLU and one dropout layer on top of the transformer model architecture for outputting the score. Regularization was also applied in form of weight decay of 0.1 and dropout of 0.1.I also used a relatively lower learning rate (1e-5) which is on the lower side of the prescribed range in the Roberta paper (1e-5,2e-5,3e-5). I did this to ensure that there is good convergence – lower the learning rate, better is the convergence for deeper networks. In the SciEntsBank graph above, we can notice that there is divergence in the second epoch. I can say that happened because we have warmup at the beginning of the training process. I used batch size of 1 as a hyperparameter which is unconventional in nature, but it provided me with the advantage of having a faster convergence.

# 6 CONCLUSION

Since the system achieved state of the art results on two public datasets accepted by the research community, I can safely say that using the SOTA transformer networks do perform better than priorly used rule-based methods. From the looks of it, training these networks ensures production ready systems which would be much tougher to break than other methods. These networks do however suffer when heavily skewed dependent variables are fed into the network (like in the case of Mohler Dataset). Yet, certain workarounds in the code can help resolve these problems. Additionally, even transformer models do have some drawbacks wherein the system could witness adversarial attacks by usage of certain keywords which have been explored in (Filighera *et al.* 2020). There can off course be workarounds for these attacks by modifying the code but such attacks do put the efficacy of system into question. For these reasons, a lot of research is going on towards using explainable deep learning-based systems which could explain what is going on in the network in real time.

# 7 EXPERIMENTS

This section discusses on various experiments I did to arrive at the results

## 7.1 MOHLER EXTENDED DATASET

### 7.1.1 FINETUNING WITH ENTIRE ARCHITECTURE FREEZED

I finetuned several transformer models on the Mohler Dataset by freezing their entire architectures. This helped save a lot of time in terms of not having to train the whole architecture to arrive at the best performing models. Finetuning on frozen transformer architecture took only 3 minutes per epoch to train while finetuning on unfrozen transformer architecture took 10 minutes per epoch to train .

This work has been a learning journey for me as I am relatively new to deep learning based architectures and hence I used RMSE loss function below while I arrived at my final results in the results section above using Huber Loss. I also used a relatively slick architecture with fewer layers on top of the transformer architecture for my final results where I had unfrozen the entire transformer architecture.

For my experiments below, I used the below hyperparameters for all models:

Batch size = 1, weight decay = 1e-8, warmup proportion = 0.06, 3 linear layers for "Large" models with one dropout (0.1) layers and one Relu layer after each linear layer except in the final layer i.e. in total seven layers. For "Base" models, I used two linear layers with dropout (0.1) and Relu layers applied after one layer. Random seed was set at 42.

Before applying any deep learning process, I scaled the score in range of 0 and 1 for feeding it into network for accurate performance.

I set batch size to 1 and not 16 or 32 (as recommended by various papers in literature) because using a batch size of 16 to 32 did not lead to any learning. Training Loss function stayed consistently stagnant during all epoch runs if I use batch size of 16 or 32. Have a look at the below line chart (Figure 19) as an example where I have used '*roberta-large-mnli*' (which generates state of the art results on ASAG in literature) with batch size of 16. This is just one example. Results have been very bad for all models I have experimented with batch size of 16/32. Note: These results were achieved by freezing the entire architecture of the transformer models. That may be the reason why the batch size of 16/32 dint show learning. By unfreezing the architecture, learning can be achieved with 16/32 batch size but in this work, I have used batch size of 1 to arrive at the results. I believe there is nothing wrong in doing this as batch size is afterall a hyperparameter and in general, it is a known fact that smaller batch sizes tend to converge better. Since our problem is a regression problem with small dataset, I believe batch size of 1 would be better to use.

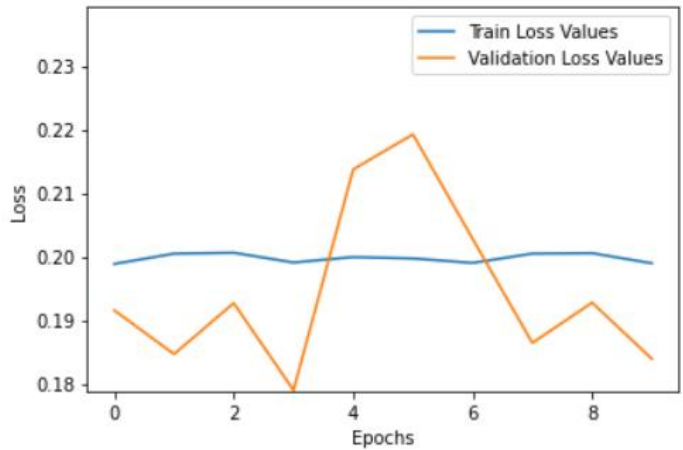

**FIGURE 19 PROBLEM WITH BATCH SIZE OF 16/32**

Upon using batch size of 1, Learning rate of 3e-5 and running for 10 epochs on the '*roberta-large-mnli*' model, got the below learning curves:

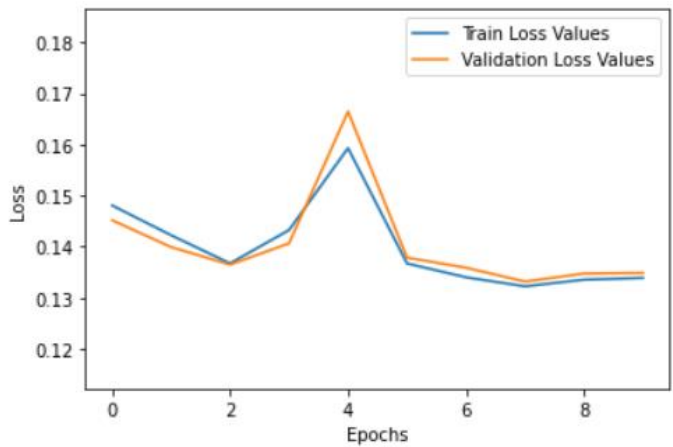

**FIGURE 20 ROBERTA-LARGE-MNLI MODEL USED FOR MOHLER**

In Figure 20, we can see that the validation loss is lower than the training loss in the first few epochs. That is because of regularization applied in form of layers (three), dropout (0.1) and weight decay (1e-8). This has been observed in even the unseen answers split of SciEntsBank dataset in this project. When I had used learning rate of 1e-5 and 2 layers for large models, the learning curves were even worser with the validation loss being lower than training loss during the entire 10 epochs:

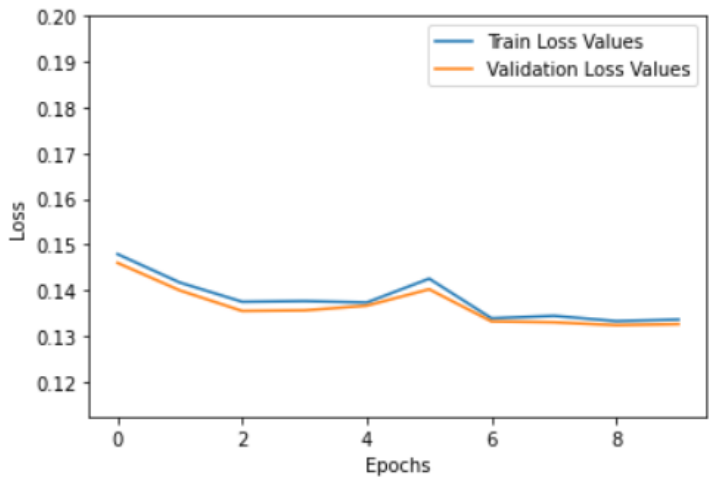

**FIGURE 21 ROBERTA-LARGE-MNLI USED ON MOHLER - LEARNING UNKNOWN**

To tackle this problem of validation loss being lower, I increased the learning rate from 1e-5 to 3e-5 and increased layers from 2 layers to 3 layers and then got the results in Figure 20.

Initially, I thought the main reason behind the validation loss being low continuously in Figure 21 is because there are duplicate student answer and reference answer pairs in the training and validation set. In machine learning terminology – this is called data leakage. This was observed in the SciEntsBank dataset Unseen Answers split as well (which I will discuss in upcoming sections). However, after removing duplicate answer pairs completely from the Mohler extended dataset, I found that the learning curves still showed that validation loss was lower than the training loss. Based on this, I understood that the reason is regularization and to a lesser extent because the answers to the same question are there in training and validation sets.

Note: After removing duplicate pairs, the performance wasn't impacted much on an aggregate basis.

One more thing that we can notice in above charts is that the losses seem to change in every epoch. I have created the system such that it would pick the epoch with the lowest validation loss during 12 epochs. I have set the system to train for 12 epochs because from literature (Camus and Filighera 2020) as well as above, I have understood that the Roberta models seem to fit well until 12 epochs.

Using 'cross-encoder/stsb-roberta-large' with learning rate of 1e-5 and 10 epochs, got the below learning curves:

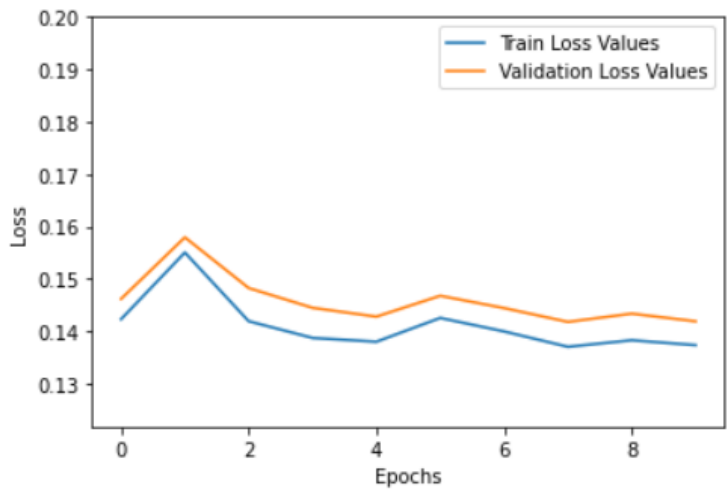

**FIGURE 22 STSB-ROBERTA-LARGE USED ON MOHLER**

Looking at the lowest level the validation loss goes to in Figure 22, it is understandable that Roberta model trained on MNLI performs better than that trained on STS corpus.

Using 'cross-encoder/quora-roberta-large' with learning rate of 1e-5 and 5 epochs, got the below learning curves:

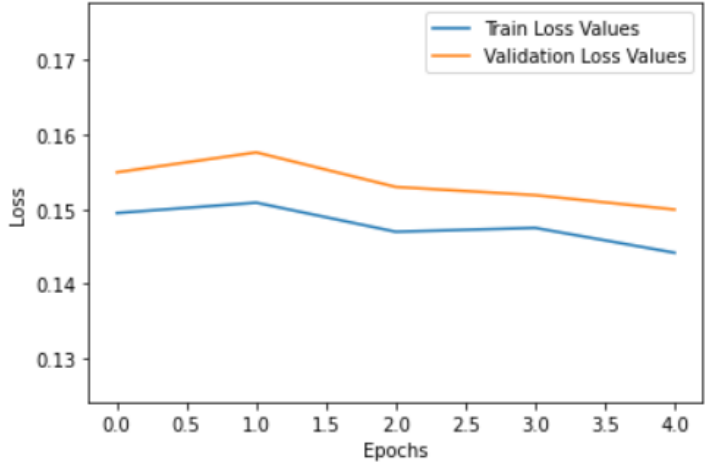

**FIGURE 23 QUORA-ROBERTA-LARGE USED ON MOHLER**

Looking at the lowest level the validation loss goes to in Figure 23, it is understandable that Roberta model trained on MNLI performs better than that trained on Quora corpus.

Using allenai/scibert_scivocab_uncased with 5 epochs and 5e-5 learning rate, got the below learning curves:

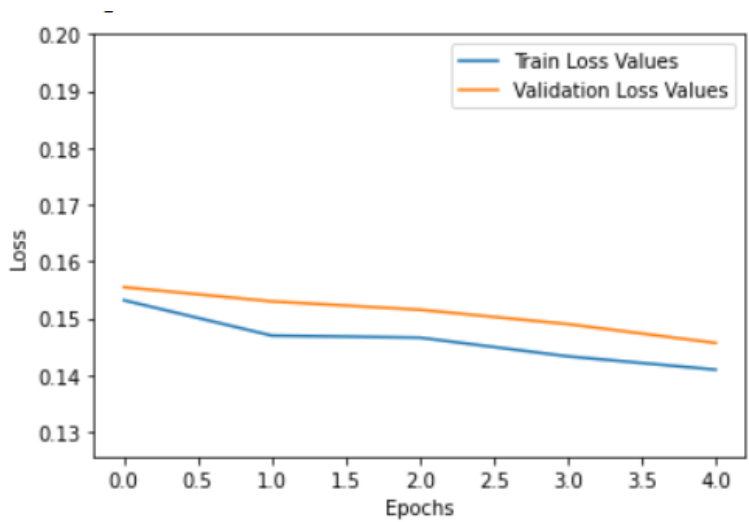

**FIGURE 24 SCIBERT_SCIVOCAB_UNCASED USED ON MOHLER**

As can be seen in Figure 24, SciBert gives a mediocre performance in terms of learning despite the dataset being related to Introductory Computer Science course.

Using 'sentence-transformers/stsb-mpnet-base-v2' with 3e-5 learning rate and 12 epochs, got the below learning curves:

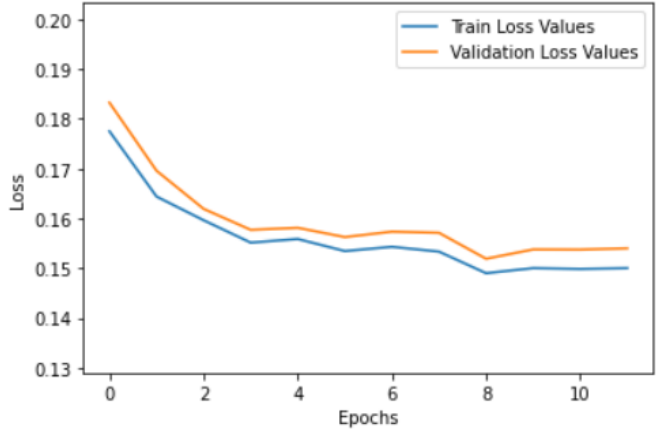

**FIGURE 25 STSB-MPNET-BASE-V2 USED ON MOHLER**

Using 'albert-large-v2' with 4 epochs and 3e-5 learning rate, got the below learning curves:

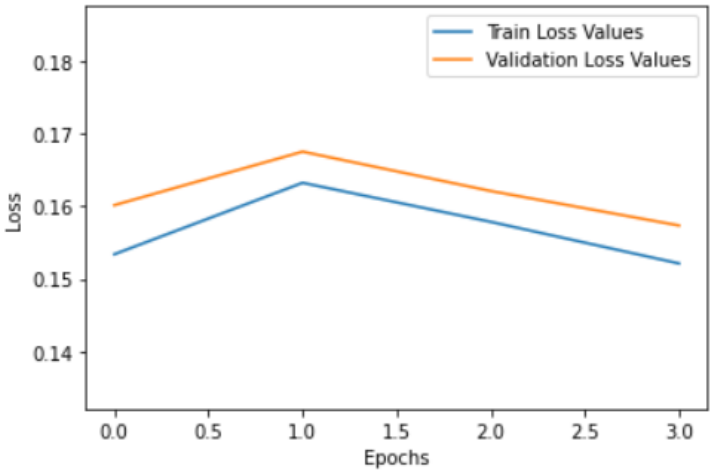

**FIGURE 26 ALBERT-LARGE-V2 USED ON MOHLER**

Using the various models discussed above, got the below performance on the test and validation set for correlation and RMSE:

**TABLE 1 FINETUNING PERFORMANCE ON MOHLER**

| Set | Correlation | RMSE |
| --- | --- | --- |
| albert-large-v2 Validation | 0.18 | 0.80 |
| albert-large-v2 Test | 0.18 | 0.85 |
| microsoft/codebert-base Validation | 0.13 | 0.80 |
| microsoft/codebert-base Test | 0.13 | 0.85 |
| cross-encoder/quora-roberta-large Validation | 0.22 | 0.75 |
| cross-encoder/quora-roberta-large Test | 0.40 | 0.75 |
| roberta-large-mnli Validation | 0.44 | **0.65** |
| roberta-large-mnli Test | 0.41 | 0.75 |
| allenai/scibert_scivocab_uncased Validation | 0.27 | 0.80 |
| allenai/scibert_scivocab_uncased Test | 0.12 | 0.85 |
| sentence-transformers/stsb-mpnet-base-v2 Validation | 0.26 | 0.75 |
| sentence-transformers/stsb-mpnet-base-v2 Test | 0.21 | 0.85 |
| cross-encoder/stsb-roberta-large Validation | 0.39 | 0.70 |
| cross-encoder/stsb-roberta-large Test | **0.51** | 0.75 |

As can be seen, generalization has been good on Roberta MNLI and Roberta STS models in Table 1. If one could closely notice, the RMSE was as low as 0.13 in some cases (which equates to 0.65 after removing scaling). The correlation however is not so good. These results however did provide a good starting point for my research as I then used the Roberta large models above for my experimenting and arriving at the final results by training the entire architecture of the transformer.

### 7.1.2 SENTENCE TRANSFORMERS

Since the models found on Huggingface Library have been pretrained on different corpuses and further finetuned on different tasks, they do hold semantic information in them for a variety of tasks. (Devlin *et al.* 2018) has a mention that BERT can be used for both finetuning and feature-based approaches. I hence decided to use the huggingface models like BERT purely for feature extraction reasons so that I could generate vectors for student answers and reference answers separately and then compute similarity metrics between the vectors in vector space. Different similarity metrics that I computed include cosine similarity, Euclidean distance, Manhattan Distance, Canberra Distance, Braycurtis dissimilarity, dice coefficient and Minkowski Distance. I also computed Skew and Kurtosis of the student answer vectors and reference answer vectors each. For feature extraction, I make use of Sentence Transformers.

- Sentence Transformers: (Reimers and Gurevych 2019) presents an alteration of the BERT network that uses Siamese and triplet network structures to derive semantically meaningful sentence embeddings. These can then be used for finding cosine similarity between the fixed size sentence vectors. Using these allows efficient feature extractions from already pretrained models rather than using other methods.

- Cosine similarity is a similarity measure between two vectors of an inner product space that measures the cosine of the angle between them. (Gomaa and Fahmy 2013)
- Euclidean Distance also called as L2 distance is the squared root of sum of squared differences between corresponding elements of two vectors. (Gomaa and Fahmy 2013)
- Manhattan Distance also called as L1 distance computes the distance that would be travelled from one data point to the other if a grid-like path would be followed. The distance is computed as the sum of the differences of their corresponding components. (Gomaa and Fahmy 2013)
- Canberra Distance is the sum of absolute values of the differences between ranks divided by their sum. It is hence also called weighted L1 distance. (Jurman *et al.* 2009)
- Minkowski Distance is a generalization of other distance measures like Hamming distance and Euclidean Distance. (Merigo and Casanovas 2011)
- Dice Coefficient is 2 into intersection between two vector sets divided by the total of the two sets. It maintains accuracy on diverse datasets and gives less weight to datasets containing unrelated features. (Henderi and Winarno 2021)
- Jaccard Distance: is 2 into intersection between two vector sets divided by union of the two sets. (Henderi and Winarno 2021)
- Braycurtis Distance: is based on braycurtis dissimilarity (Bray and Curtis 1957) used in ecology and used in quantifying compositional dissimilarity between two different sites.
- Skew and Kurtosis of Student Answers and Reference Answers Each

Got the below results by using various Similarity Measures between vectors generated by SBERT as features fed into ScikitLearn's RandomForestRegressor with default parameters:

**TABLE 2 RANDOMFORESTREGRESSOR RESULTS ON SBERT ON MOHLER**

| Metric | Result |
| --- | --- |
| Mean Absolute Error | 0.60 |
| Root Mean Squared Error: | 0.80 |
| R-Squared: | 0.53 |
| Correlation: | 0.73 |

As can be seen, SBERT seems to singlehandedly produce great results on the dataset (SOTA Correlation is 0.80).

### 7.1.3 UNIVERSAL SENTENCE ENCODERS DEEP AVERAGING NETWORK (USE DAN)

USE (Cer *et al.* 2018) provides a way to generate sentence embeddings using transfer learning. I used vector representations of student and reference answers from USE and found cosine similarities and other similarity metrics between vectors.

Got the below results by using various Similarity Measures between vectors generated by USE as features fed into ScikitLearn's RandomForestRegressor with default parameters:

**TABLE 3 RANDOMFORESTREGRESSOR RESULTS ON USE MOHLER**

| Metric | Result |
| --- | --- |
| Mean Absolute Error | 0.75 |
| Root Mean Squared Error: | 1.00 |
| R-Squared: | 0.24 |
| Correlation: | 0.49 |

### 7.1.4 HANDCRAFTED FEATURES AND FUZZYWUZZY FEATURES

I also experimented with using Other Handcrafted and FuzzyWuzzy Metrics (Varma 2018) as features like common word count by minimum of Student Answer and Reference answer words length, common word count by maximum of student answer and reference answer words length, common stop word count by minimum of student and reference answer stop words length, common stop word count by maximum of student and reference answer stop words length, common token count by minimum of student and reference answer token length, common token count by maximum of student and reference answer token length, common last word represented as a one-hot encoded variable, common first word represented as a one-hot encoded variable, difference between number of tokens in the two answers and average token length of the two answers. Prior to computing these, stemming was performed on the answers for better performance.

Got the below results by using Handcrafted Features and fuzzywuzzy features:

**TABLE 4 RANDOMFOREST REGRESSOR RESULTS ON HANDCRAFTED AND FUZZYWUZZY FEATURES ON MOHLER**

| Metric | Result |
| --- | --- |
| Mean Absolute Error | 0.70 |
| Root Mean Squared Error: | 0.85 |
| R-Squared: | 0.32 |

| Correlation: | 0.57 |

## 7.2 SCIENTSBANK (SEB) DATASET

### 7.2.1 FINETUNING BY FREEZING THE ARCHITECTURE

I used the below parameters for finetuning all the models for this dataset:

Batch size = 1

Weight decay = 1e-8

Warmup proportion = 0.06

Using Learning rate of 3e-5 with 12 epochs, got the below performance from the finetuning *roberta-large-mnli* model:

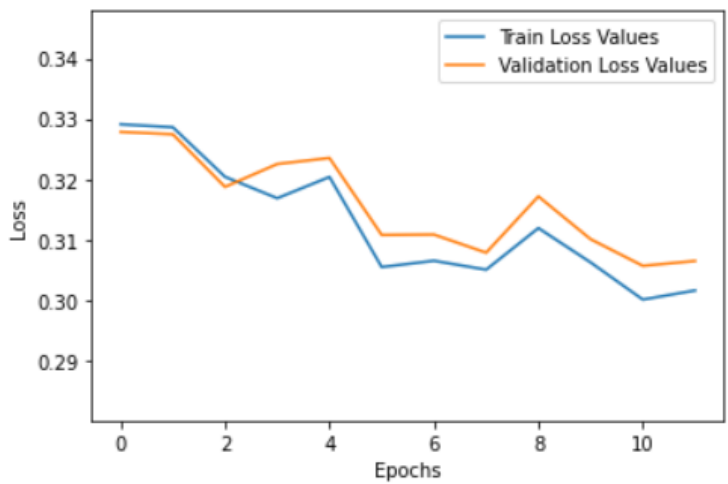

**FIGURE 27 ROBERTA LARGE MNLI FINETUNED ON SEB**

Note: As per the original Roberta paper (Y. Liu *et al.* 2019), it is recommended to use 10 epochs or less for Roberta models. However, I used 12 epochs because they had explored using more than 10 epochs in this paper - (Camus and Filighera 2020). They found that the same performance was observed in 12 epochs or more for this dataset.

Using cross-encoder/stsb-roberta-large with learning rate of 1e-5 and 8 epochs, got the below performance:

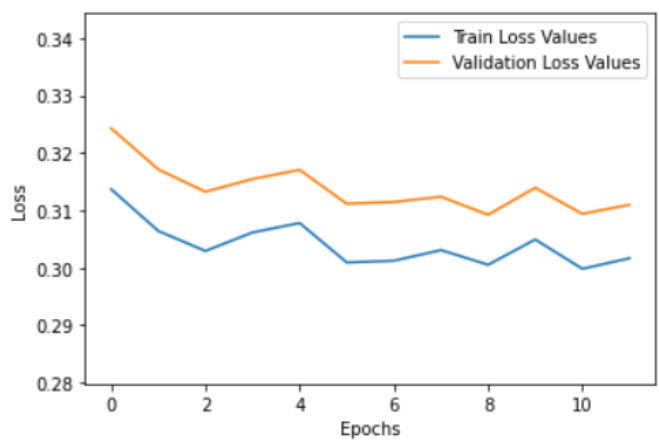

**FIGURE 28 STSB-ROBERTA-LARGE FINETUNED ON SEB**

Using the above two models, got the below results on Random Forest Regressor default parameters:

**TABLE 5 RAMDOMFORESTREGRESSOR RESULTS USING FINETUNED MODELS ON SEB**

| Metric | Result |
| --- | --- |
| Mean Absolute Error | 0.60 |
| Root Mean Squared Error: | 0.76 |
| R-Squared: | 0.26 |
| Correlation: | 0.50 |

### 7.2.2 SENTENCE TRANSFORMERS

Got the below results by using various Similarity Measures between vectors generated by SBERT as features fed into ScikitLearn's RandomForestRegressor with default parameters:

**TABLE 6 RANDOMFORESTREGRESSOR RESULTS ON SBERT SEB**

| Metric | Result |
| --- | --- |
| Mean Absolute Error | 0.56 |
| Root Mean Squared Error: | 0.66 |
| R-Squared: | 0.43 |
| Correlation: | 0.66 |

### 7.2.3 UNIVERSAL SENTENCE ENCODERS DEEP AVERAGING NETWORK (USE DAN)

Got the below results by using various Similarity Measures between vectors generated by USE DAN as features fed into ScikitLearn's RandomForestRegressor with default parameters:

**TABLE 7 RANDOMFORESTREGRESSOR RESULTS ON USE SEB**

| Metric | Result |
|---|---|
| Mean Absolute Error | 0.68 |
| Root Mean Squared Error: | 0.82 |
| R-Squared: | 0.14 |
| Correlation: | 0.37 |

### 7.2.4 HANDCRAFTED FEATURES AND FUZZYWUZZY RATIOS

Got the below results by using handcrafted features and FuzzyWuzzy Ratios as features fed into ScikitLearn's RandomForestRegressor with default parameters:

**TABLE 8 RANDOMFORESTREGRESSOR RESULTS ON HANDCRAFTED FEATURES AND FUZZYWUZZY RATIOS SEB**

| Metric | Result |
|---|---|
| Mean Absolute Error | 0.66 |
| Root Mean Squared Error: | 0.78 |
| R-Squared: | 0.24 |
| Correlation: | 0.49 |

**Reference List**

Agirre, E., Banea, C., Cardie, C., Cer, D., Diab, M., Gonzalez-Agirre, A., Guo, W., Lopez-Gazpio, I., Maritxalar, M. and Mihalcea, R. (2015) 'Semeval-2015 task 2: Semantic textual similarity, english, spanish and pilot on interpretability', in *Proceedings of the 9th international workshop on semantic evaluation (SemEval 2015)*, 252-263.

Agirre, E., Banea, C., Cardie, C., Cer, D., Diab, M., Gonzalez-Agirre, A., Guo, W., Mihalcea, R., Rigau, G. and Wiebe, J. (2014) 'Semeval-2014 task 10: Multilingual semantic textual similarity', in *Proceedings of the 8th international workshop on semantic evaluation (SemEval 2014)*, 81-91.

Agirre, E., Cer, D., Diab, M. and Gonzalez-Agirre, A. (2012) 'Semeval-2012 task 6: A pilot on semantic textual similarity', in *\* SEM 2012: The First Joint Conference on Lexical and Computational Semantics–Volume 1: Proceedings of the main conference and the shared task, and Volume 2: Proceedings of the Sixth International Workshop on Semantic Evaluation (SemEval 2012)*, 385-393.

Agirre, E., Cer, D., Diab, M., Gonzalez-Agirre, A. and Guo, W. (2013) '\* SEM 2013 shared task: Semantic textual similarity', in *Second joint conference on lexical and computational semantics (\* SEM),* *volume 1: proceedings of the Main conference and the shared task: semantic textual similarity*, 32-43.

Basak, R., Naskar, S.K. and Gelbukh, A. (2019) 'Short-answer grading using textual entailment', *Journal of Intelligent & Fuzzy Systems*, 36(5), 4909-4919.

Basu, S., Jacobs, C. and Vanderwende, L. (2013) 'Powergrading: a clustering approach to amplify human effort for short answer grading', *Transactions of the Association for Computational Linguistics*, 1, 391-402.

Beltagy, I., Lo, K. and Cohan, A. (2019) 'Scibert: A pretrained language model for scientific text', *arXiv preprint arXiv:1903.10676*.

Bray, J. and Curtis, J. (1957) 'An ordination of upland forest communities of southern Wisconsin. Ecological Monographs (27)', 325-349.

Brooks, M., Basu, S., Jacobs, C. and Vanderwende, L. (2014) 'Divide and correct: Using clusters to grade short answers at scale', in *Proceedings of the first ACM conference on Learning@ scale conference*, 89-98.

Brown, T.B., Mann, B., Ryder, N., Subbiah, M., Kaplan, J., Dhariwal, P., Neelakantan, A., Shyam, P., Sastry, G. and Askell, A. (2020) 'Language models are few-shot learners', *arXiv preprint arXiv:2005.14165*.

Burrows, S., Gurevych, I. and Stein, B. (2015) 'The eras and trends of automatic short answer grading', *International Journal of Artificial Intelligence in Education, 25(1)*, 60-117.

Camus, L. and Filighera, A. (2020) 'Investigating transformers for automatic short answer grading', in *International Conference on Artificial Intelligence in Education*, Springer, 43-48.

Cer, D., Yang, Y., Kong, S.-y., Hua, N., Limtiaco, N., John, R.S., Constant, N., Guajardo-Céspedes, M., Yuan, S. and Tar, C. (2018) 'Universal sentence encoder', *arXiv preprint arXiv:1803.11175*.

Chaturvedi, B. and Basak, R. (2021) 'Automatic Short Answer Grading Using Corpus-Based Semantic Similarity Measurements' in *Progress in Advanced Computing and Intelligent Engineering* Springer, 266-281.

Condor, A. (2020) 'Exploring Automatic Short Answer Grading as a Tool to Assist in Human Rating', in *International Conference on Artificial Intelligence in Education*, Springer, 74-79.

Cutrone, L. and Chang, M. (2011) 'Auto-assessor: computerized assessment system for marking student's short-answers automatically', in *2011 IEEE International Conference on Technology for Education*, IEEE, 81-88.

Devlin, J., Chang, M.-W., Lee, K. and Toutanova, K. (2018) 'Bert: Pre-training of deep bidirectional transformers for language understanding', *arXiv preprint arXiv:1810.04805*.

Dzikovska, M.O., Nielsen, R. and Brew, C. (2012) 'Towards effective tutorial feedback for explanation questions: A dataset and baselines', in *Proceedings of the 2012 Conference of the North American Chapter of the Association for Computational Linguistics: Human Language Technologies*, 200-210.

Dzikovska, M.O., Nielsen, R.D., Brew, C., Leacock, C., Giampiccolo, D., Bentivogli, L., Clark, P., Dagan, I. and Dang, H.T. (2013) *Semeval-2013 task 7: The joint student response analysis and 8th recognizing textual entailment challenge*: NORTH TEXAS STATE UNIV DENTON.

Dzikovska, M.O., Nielsen, R.D. and Leacock, C. (2016) 'The joint student response analysis and recognizing textual entailment challenge: making sense of student responses in educational applications', *Language Resources and Evaluation*, 50(1), 67-93.

Feng, Z., Guo, D., Tang, D., Duan, N., Feng, X., Gong, M., Shou, L., Qin, B., Liu, T. and Jiang, D. (2020) 'Codebert: A pre-trained model for programming and natural languages', *arXiv preprint arXiv:2002.08155*.

Filighera, A., Steuer, T. and Rensing, C. (2020) 'Fooling automatic short answer grading systems', in *International Conference on Artificial Intelligence in Education*, Springer, 177-190.

Galhardi, L., Barbosa, C.R., de Souza, R.C.T. and Brancher, J.D. (2018) 'Portuguese automatic short answer grading', in *Brazilian Symposium on Computers in Education (Simpósio Brasileiro de Informática na Educação-SBIE)*, 1373.

Galhardi, L.B. and Brancher, J.D. (2018) 'Machine learning approach for automatic short answer grading: A systematic review', *In Ibero-american conference on artificial intelligence (pp. 380-391). Springer, Cham*.

Ghavidel, H.A., Zouaq, A. and Desmarais, M.C. (2020) 'Using BERT and XLNET for the Automatic Short Answer Grading Task', in *CSEDU (1)*, 58-67.

Gomaa, W.H. and Fahmy, A.A. (2012) 'Short answer grading using string similarity and corpus-based similarity', *International Journal of Advanced Computer Science and Applications (IJACSA)*, 3(11).

Gomaa, W.H. and Fahmy, A.A. (2013) 'A survey of text similarity approaches', *International Journal of Computer Applications*, 68(13), 13-18.

Gomaa, W.H. and Fahmy, A.A. (2014) 'Arabic short answer scoring with effective feedback for students', *International Journal of Computer Applications*, 86(2), 35-41.

Gomaa, W.H. and Fahmy, A.A. (2019) 'Ans2vec: A scoring system for short answers', in *International Conference on Advanced Machine Learning Technologies and Applications*, Springer, 586-595.

Hassan, S., Fahmy, A.A. and El-Ramly, M. (2018) 'Automatic short answer scoring based on paragraph embeddings', *International Journal of Advanced Computer Science and Applications*, 9(10), 397-402.

Henderi, H. and Winarno, W. (2021) 'Text Mining an Automatic Short Answer Grading (ASAG), Comparison of Three Methods of Cosine Similarity, Jaccard Similarity and Dice's Coefficient', *Journal of Applied Data Sciences*, 2(2).

Hewlett, T.W.a.F. (2012) *The Hewlett Foundation: Short Answer Scoring* [dataset], available: https://www.kaggle.com/c/asap-sas/overview.

Horbach, A. and Pinkal, M. (2018) 'Semi-supervised clustering for short answer scoring', in *Proceedings of the Eleventh International Conference on Language Resources and Evaluation (LREC 2018)*.

Huber, P.J. (1992) 'Robust estimation of a location parameter' in *Breakthroughs in statistics* Springer, 492-518.

Jang, E.-S., Kang, S.-S., Noh, E.-H., Kim, M.-H., Sung, K.-H. and Seong, T.-J. (2014) 'KASS: Korean Automatic Scoring System for Short-answer Questions', in *CSEDU (2)*, 226-230.

Jayashankar, S. and Sridaran, R. (2017) 'Superlative model using word cloud for short answers evaluation in eLearning', *Education and Information Technologies*, 22(5), 2383-2402.

Jurman, G., Riccadonna, S., Visintainer, R. and Furlanello, C. (2009) 'Canberra distance on ranked lists', in *Proceedings of advances in ranking NIPS 09 workshop*, Citeseer, 22-27.

Kishaan, J., Muthuraja, M., Nair, D. and Plöger, P.G. (2020) 'Using Active Learning for Assisted Short Answer Grading'.

Klein, R., Kyrilov, A. and Tokman, M. (2011) 'Automated assessment of short free-text responses in computer science using latent semantic analysis', in *Proceedings of the 16th annual joint conference on Innovation and technology in computer science education*, 158-162.

Kulkarni, C.E., Socher, R., Bernstein, M.S. and Klemmer, S.R. (2014) 'Scaling short-answer grading by combining peer assessment with algorithmic scoring', in *Proceedings of the first ACM conference on Learning@ scale conference*, 99-108.

Kumar, S., Chakrabarti, S. and Roy, S. (2017) 'Earth Mover's Distance Pooling over Siamese LSTMs for Automatic Short Answer Grading', in *IJCAI*, 2046-2052.

Lan, Z., Chen, M., Goodman, S., Gimpel, K., Sharma, P. and Soricut, R. (2019) 'Albert: A lite bert for self-supervised learning of language representations', *arXiv preprint arXiv:1909.11942*.

Liu, T., Ding, W., Wang, Z., Tang, J., Huang, G.Y. and Liu, Z. (2019) 'Automatic short answer grading via multiway attention networks', in *International conference on artificial intelligence in education*, Springer, 169-173.

Liu, Y., Ott, M., Goyal, N., Du, J., Joshi, M., Chen, D., Levy, O., Lewis, M., Zettlemoyer, L. and Stoyanov, V. (2019) 'Roberta: A robustly optimized bert pretraining approach', *arXiv preprint arXiv:1907.11692*.

Lun, J., Zhu, J., Tang, Y. and Yang, M. (2020) 'Multiple data augmentation strategies for improving performance on automatic short answer scoring', in *Proceedings of the AAAI Conference on Artificial Intelligence*, 13389-13396.

Magooda, A.E., Zahran, M., Rashwan, M., Raafat, H. and Fayek, M. (2016) 'Vector based techniques for short answer grading', in *The twenty-ninth international flairs conference*.

Merigo, J.M. and Casanovas, M. (2011) 'A new Minkowski distance based on induced aggregation operators', *International Journal of Computational Intelligence Systems*, 4(2), 123-133.

Meurers, D., Ziai, R., Ott, N. and Bailey, S.M. (2011a) 'Integrating parallel analysis modules to evaluate the meaning of answers to reading comprehension questions', *International Journal of Continuing Engineering Education and Life Long Learning*, 21(4), 355-369.

Meurers, D., Ziai, R., Ott, N. and Kopp, J. (2011b) 'Evaluating answers to reading comprehension questions in context: Results for German and the role of information structure', in *Proceedings of the TextInfer 2011 Workshop on Textual Entailment*, 1-9.

Mohler, M., Bunescu, R. and Mihalcea, R. (2011) 'Learning to grade short answer questions using semantic similarity measures and dependency graph alignments', in *Proceedings of the 49th annual meeting of the association for computational linguistics: Human language technologies*, 752-762.

Mohler, M., Bunescu, R. and Mihalcea, R. (2011) *A larger collection of short student answers and grades for a course in Computer Science* [dataset], available: https://web.eecs.umich.edu/~mihalcea/downloads.html.

Mohler, M. and Mihalcea, R. (2009) 'Text-to-text semantic similarity for automatic short answer grading', in *Proceedings of the 12th Conference of the European Chapter of the ACL (EACL 2009)*, 567-575.

Nielsen, R.D., Ward, W.H., Martin, J.H. and Palmer, M. (2008) 'Annotating Students' Understanding of Science Concepts', in *LREC*, Citeseer.

Omran, A.M.B. and Ab Aziz, M.J. (2013) 'Automatic essay grading system for short answers in English language', *Journal of Computer Science*, 9(10), 1369.

Omran, B., Muftah, A., Aziz, A. and Juzaiddin, M. (2014) 'Syntactically enhanced LSA methods in automatic essay grading systems for short answers', in *Proceedings of the 3rd International Conference on Computer Engineering and Mathematical Sciences (ICCEMS 2014)*, 412-417.

Pado, U. and Kiefer, C. (2015) 'Short answer grading: When sorting helps and when it doesn't', in *Proceedings of the fourth workshop on NLP for computer-assisted language learning*, 42-50.

Pribadi, F.S., Permanasari, A.E. and Adji, T.B. (2018) 'Short answer scoring system using automatic reference answer generation and geometric average normalized-longest common subsequence (GAN-LCS)', *Education and Information Technologies*, 23(6), 2855-2866.

Radford, A., Narasimhan, K., Salimans, T. and Sutskever, I. (2018) 'Improving language understanding by generative pre-training'.

Radford, A., Wu, J., Child, R., Luan, D., Amodei, D. and Sutskever, I. (2019) 'Language models are unsupervised multitask learners', *OpenAI blog*, 1(8), 9.

Reimers, N. and Gurevych, I. (2019) 'Sentence-bert: Sentence embeddings using siamese bert-networks', *arXiv preprint arXiv:1908.10084*.

Riordan, B., Horbach, A., Cahill, A., Zesch, T. and Lee, C. (2017) 'Investigating neural architectures for short answer scoring', in *Proceedings of the 12th Workshop on Innovative Use of NLP for Building Educational Applications*, 159-168.

Roy, D., Paul, D., Mitra, M. and Garain, U. (2016) 'Using word embeddings for automatic query expansion', *arXiv preprint arXiv:1606.07608*.

Roy, S., Dandapat, S., Nagesh, A. and Narahari, Y. (2016) 'Wisdom of students: A consistent automatic short answer grading technique', in *Proceedings of the 13th International Conference on Natural Language Processing*, 178-187.

Sadr, H. and Nazari Solimandarabi, M. (2019) 'Presentation of an efficient automatic short answer grading model based on combination of pseudo relevance feedback and semantic relatedness measures', *Journal of Advances in Computer Research*, 10(2), 17-30.

Saha, S., Dhamecha, T.I., Marvaniya, S., Sindhgatta, R. and Sengupta, B. (2018) 'Sentence level or token level features for automatic short answer grading?: Use both', in *International conference on artificial intelligence in education*, Springer, 503-517.

Sahu, A. and Bhowmick, P.K. (2019) 'Feature engineering and ensemble-based approach for improving automatic short-answer grading performance', *IEEE Transactions on Learning Technologies*, 13(1), 77-90.

Sil, A., Shelton, A., Ketelhut, D.J. and Yates, A. (2012) 'Automatic grading of scientific inquiry', in *Proceedings of the Seventh Workshop on Building Educational Applications Using NLP*, 22-32.

Song, K., Tan, X., Qin, T., Lu, J. and Liu, T.-Y. (2020) 'Mpnet: Masked and permuted pre-training for language understanding', *arXiv preprint arXiv:2004.09297*.

Sultan, M.A., Bethard, S. and Sumner, T. (2015) 'Dls@ cu: Sentence similarity from word alignment and semantic vector composition', in *Proceedings of the 9th International Workshop on Semantic Evaluation (SemEval 2015)*, 148-153.

Sultan, M.A., Bethard, S. and Sumner, T. (2016a) 'DLS@ CU at Semeval-2016 Task 1: Supervised models of sentence similarity', in *Proceedings of the 10th International Workshop on Semantic Evaluation (SemEval-2016)*, 650-655.

Sultan, M.A., Salazar, C. and Sumner, T. (2016b) 'Fast and easy short answer grading with high accuracy', in *Proceedings of the 2016 Conference of the North American Chapter of the Association for Computational Linguistics: Human Language Technologies*, 1070-1075.

Sung, C, Dhamecha, T., Saha, S., Ma, T., Reddy, V. and Arora, R. (2019a) 'Pre-training BERT on domain resources for short answer grading', in *Proceedings of the 2019 Conference on Empirical Methods in Natural Language Processing and the 9th International Joint Conference on Natural Language Processing (EMNLP-IJCNLP)*, 6071-6075.

Sung, C., Dhamecha, T.I. and Mukhi, N. (2019b) 'Improving short answer grading using transformer-based pre-training', in *International Conference on Artificial Intelligence in Education*, Springer, 469-481.

Surya, K., Gayakwad, E. and Nallakaruppan, M. (2019) 'Deep learning for short answer scoring', *Int. J. Recent Technol. Eng*, 7(6), 1712-1715.

Süzen, N., Gorban, A.N., Levesley, J. and Mirkes, E.M. (2020) 'Automatic short answer grading and feedback using text mining methods', *Procedia Computer Science*, 169, 726-743.

Tan, H., Wang, C., Duan, Q., Lu, Y., Zhang, H. and Li, R. (2020) 'Automatic short answer grading by encoding student responses via a graph convolutional network', *Interactive learning environments*, 1-15.

Tulu, C.N., Ozkaya, O. and Orhan, U. (2021) 'Automatic Short Answer Grading With SemSpace Sense Vectors and MaLSTM', *IEEE Access*, 9, 19270-19280.

Varma, S. (2018) *Case Study 1: Quora question Pair Similarity Problem*, 6: Machine Learning Real-World Case Studies, available: https://www.appliedaicourse.com/lecture/11/applied-machine-learning-online-course/4153/how-to-optimally-learn-from-case-studies-in-the-course/7/module-6-machine-learning-real-world-case-studies [accessed 4th August 2021].

Vaswani, A., Shazeer, N., Parmar, N., Uszkoreit, J., Jones, L., Gomez, A.N., Kaiser, Ł. and Polosukhin, I. (2017) 'Attention is all you need', in *Advances in neural information processing systems*, 5998-6008.

Wolf, T., Debut, L., Sanh, V., Chaumond, J., Delangue, C., Moi, A., Cistac, P., Rault, T., Louf, R. and Funtowicz, M. (2019) 'Huggingface's transformers: State-of-the-art natural language processing', *arXiv preprint arXiv:1910.03771*.

Yang, Z., Dai, Z., Yang, Y., Carbonell, J., Salakhutdinov, R.R. and Le, Q.V. (2019) 'Xlnet: Generalized autoregressive pretraining for language understanding', *Advances in neural information processing systems*, 32.

Zehner, F., Sälzer, C. and Goldhammer, F. (2016) 'Automatic coding of short text responses via clustering in educational assessment', *Educational and psychological measurement*, 76(2), 280-303.

Zhang, L., Huang, Y., Yang, X., Yu, S. and Zhuang, F. (2019) 'An automatic short-answer grading model for semi-open-ended questions', *Interactive learning environments*, 1-14.

Zhang, Y., Lin, C. and Chi, M. (2020) 'Going deeper: Automatic short-answer grading by combining student and question models', *User Modeling and User-Adapted Interaction*, 30(1), 51-80.

Zhang, Y., Shah, R. and Chi, M. (2016) 'Deep Learning+ Student Modeling+ Clustering: A Recipe for Effective Automatic Short Answer Grading', *International Educational Data Mining Society*.